\documentclass[10pt,journal,compsoc]{IEEEtran}
\usepackage{url}
\usepackage{subfigure}
\usepackage{multirow}
\usepackage{graphicx}
\usepackage{booktabs}
\usepackage{amsmath}
\usepackage{amsfonts}
\usepackage{dsfont}
\usepackage{bbm}
\usepackage{makecell}
\usepackage{color}

\ifCLASSOPTIONcompsoc
  \usepackage[nocompress]{cite}
\else
  \usepackage{cite}
\fi
\ifCLASSINFOpdf
\else
\fi
\usepackage{algorithmic}
\usepackage[ruled, linesnumbered]{algorithm2e}
\begin{document}
%
\title{Weak Supervision with Arbitrary Single Frame for Micro- and Macro-expression Spotting}
%
%
%
%

\author{Wang-Wang~Yu, Xian-Shi Zhang, Fu-Ya Luo, Yijun Cao, Kai-Fu Yang, Hong-Mei Yan, and ~Yong-Jie Li,~\IEEEmembership{Member,~IEEE}
\IEEEcompsocitemizethanks{
\IEEEcompsocthanksitem W. Yu, F. Luo, Y. Cao, K. Yang, H. Yan and Y. Li are with MOE Key Lab for Neuroinformation, 
University Of Electronic Science And Technology Of China, Chengdu,
China. Email: yuwangwang91@163.com, yangkf@uestc.edu.cn, hmyan@uestc.edu.cn, liyj@uestc.edu.cn\protect}
\thanks{Manuscript received April 19, 2005; revised August 26, 2015. (Corresponding author: Yong-Jie Li, Xian-Shi Zhang.)}}

%
%

\markboth{Journal of \LaTeX\ Class Files,~Vol.~14, No.~8, August~2015}%
{Shell \MakeLowercase{\textit{et al }}: Bare Demo of IEEEtran.cls for Computer Society Journals}
%



\IEEEtitleabstractindextext{%
\begin{abstract}
Frame-level micro- and macro-expression spotting methods require time-consuming frame-by-frame observation during annotation. Meanwhile, 
video-level spotting lacks sufficient information about the location and number of expressions during training, resulting in significantly inferior 
performance compared with fully-supervised spotting. To bridge this gap, we propose a point-level weakly-supervised expression spotting (PWES) framework, 
where each expression requires to be annotated with only one random frame (i.e., a point). To mitigate the issue of sparse label distribution, 
the prevailing solution is pseudo-label mining, which, however, introduces new problems: localizing contextual background snippets results in 
inaccurate boundaries and discarding foreground snippets leads to fragmentary predictions. Therefore, we design the strategies of multi-refined 
pseudo label generation (MPLG) and distribution-guided feature contrastive learning (DFCL) to address these problems. Specifically, MPLG generates more reliable 
pseudo labels by merging class-specific probabilities, attention scores, fused features, and point-level labels. DFCL is utilized to enhance feature 
similarity for the same categories and feature variability for different categories while capturing global representations across the entire datasets. 
Extensive experiments on the CAS(ME)$^2$, CAS(ME)$^3$, and SAMM-LV datasets demonstrate PWES achieves promising performance comparable to that of recent 
fully-supervised methods.
\end{abstract}

\begin{IEEEkeywords}
  Micro- and macro-expression spotting, point-level supervision, multi-refined pseudo label generation, 
  distribution-guided feature contrastive learning.
\end{IEEEkeywords}}

\maketitle

\IEEEdisplaynontitleabstractindextext

%
\IEEEpeerreviewmaketitle

\IEEEraisesectionheading{\section{Introduction}\label{sec:introduction}}

%
%
%
%

\IEEEPARstart{F}{acial} expressions, as a form of non-verbal communication, play a crucial role in conveying intentions and 
attitudes during interaction. They also provide intrinsic non-emotional cues, such as interest and attention. The expressions 
can be grouped into micro-expressions (MEs) and macro-expressions (MaEs) \cite{ekman1969nonverbal}. 
MEs are subtle, unconscious facial movements that typically last less than 0.5 second, while MaEs are distinctive facial 
movements that normally last from 0.5 to 4.0 seconds \cite{ekman2003darwin}. While MaEs may convey 
incorrect emotions, MEs can reveal real emotions when an individual attempts to conceal their actual emotions \cite{ekman2003darwin, 
ekman2009lie}. Hence, analyzing MEs is particularly valuable in high-stakes situations such as business negotiation, judicial practice, 
clinical diagnosis, and public safety \cite{ekman2003darwin}.

\begin{figure}[tbp]
  \centering
  \includegraphics[width=\linewidth]{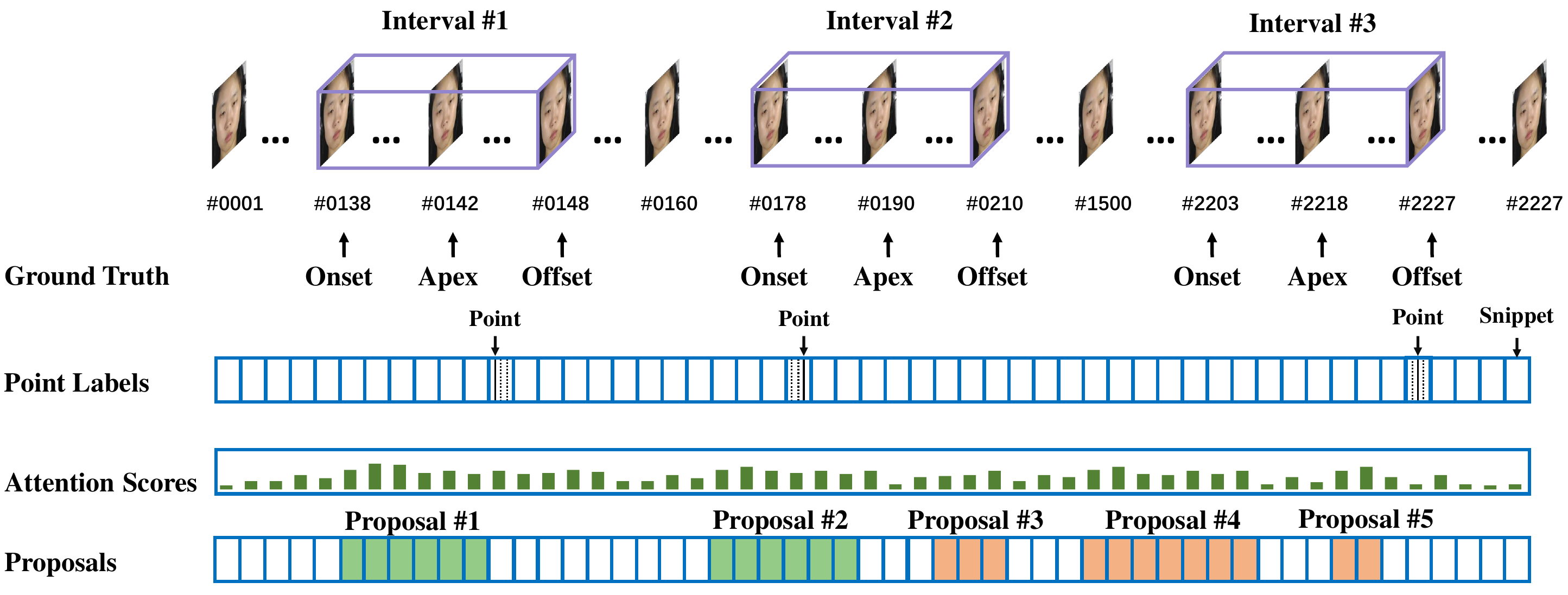}
  \caption{\textcolor{black}{A video containing frames \#1 to \#2273 from the CAS(ME)$^2$ dataset. The video contains three ground truth intervals, with the first 
  interval containing a ME and the last two containing MaEs. We first preprocess the video by dividing it into uniform, non-overlapping snippets, each of which
  contains the same number of frames. A random frame is selected from each ground truth interval as one of the point-level labels to train our model. 
  During training, we generate attention scores which signify the probabilities of foreground. During testing, we use these attention 
  scores to generate proposals with different top-$k$ values \cite{yu2023weaklysupervised}. Specifically, the green and orange blocks represent 
  valid and invalid proposal intervals, respectively. Our objective is to identify consecutive video snippets that closely match the ground truth intervals.
  }}
  \label{fig1}
\end{figure}  

The processing of expressions can be divided into two primary stages: spotting and recognition. Specifically, the recognition 
stage performs the classification of facial expressions into specific emotional categories \cite{xie2019adaptive} or multidimensional 
continuous values \cite{du2014compound, li2017reliable, liang2020fine} based on various datasets \cite{davison2016samm, li2013spontaneous,
yan2013casme, yan2014casme, zhao2011facial, kossaifi2019sewa, kollias2020analysing}. In contrast, the spotting stage serves as 
a preceding task to localize specific intervals in an untrimmed video and classify them into MEs and MaEs based on the spotting 
datasets, e.g., CAS(ME)$^2$ \cite{qu2017cas}, CAS(ME)$^3$ \cite{li2022cas}, SAMM-LV \cite{yap2020samm}, and MMEW \cite{ben2021video}. 
Considering that the main differences between MEs and MaEs lie in their duration, intensity of movement, and range of movement \cite{wang2021mesnet}, 
and that the latter two are difficult to quantify \cite{lu2022more}, the above spotting datasets use duration as the primary benchmark to quantify 
the differences between MEs and MaEs. Moreover, within the entire duration, each expression can further be characterized by the frames of onset, apex, 
and offset \cite{wang2021mesnet}. The onset and offset signify respectively the starting and ending time, and the apex reflects the most noticeable 
emotional information with maximum facial muscle deformation \cite{ekman1993facial, esposito2007amount}. The above spotting datasets provide the onset 
and offset frames of all ground truth intervals.

Consequently, the existing spotting models on the above labeled dataset can be categorized into two main groups based on their supervisory approaches: 
frame-level labeling-based methods \cite{he2022micro, yu2021lssnet, guo2023micro, yu2023lgsnet} and video-level labeling-based methods \cite{yu2023weaklysupervised}. 
Specifically, frame-level labeling-based methods are those spotting techniques 
that rely on specific frames within a video or a sequence of images as training labels. In contrast, video-level labeling-based methods use binary labels 
that indicate the presence or absence of all categories during training, without considering the quantity and locations of each category at 
particular timestamps within a video or sequence of images. 

Frame-level labeling in a fully-supervised setting has been successfully implemented in untrimmed videos for ME and MaE spotting 
\cite{he2022micro, yu2021lssnet, guo2023micro, yu2023lgsnet}. However, fine-grained frame-level labeling requires heavy manual labor by multiple coders 
through frame-by-frame observation \cite{qu2017cas, yap2020samm, ben2021video, li2022cas}. 
To address this problem, we have attempted to utilize video-level labeling to achieve ME and MaE spotting \cite{yu2023weaklysupervised}, which 
is a weakly-supervised temporal action localization (WTAL) task of facial video understanding. In the task of WTAL \cite{islam2021hybrid, hong2021cross, 
paul2018w}, video-level labels serve as weak supervision for model's training with a lack of information regarding action locations and the number of 
action instances. This is quite challenging and always results in poor performance compared to fully-supervised methods \cite{wang2017untrimmednets, 
nguyen2018weakly, nguyen2019weakly, paul2018w, su2018cascaded}. Therefore, several methods \cite{ma2020sf, lee2021learning} suggest adding point-level 
information into weakly-supervised settings by providing one labelled frame for each ground truth interval, which does not increase much cost compared 
with video-level labeling. Specifically, considering that point-level labels need more localization-related and classification-related information, apex 
frames of ground truths seem to be the ideal choice for point-level weakly-supervised expression spotting. However, such annotation is 
still dependent on frame-by-frame observation \cite{li2022cas}. To effectively reduce annotation cost, we propose to use one arbitrary frame within each 
ground truth interval as its point-level label, which is regarded as a coarse cue to help localize key snippets during training. As shown in 
Figure \ref{fig1}, we utilize arbitrary single frame as weak supervision information to construct our point-level framework by compensating for the insufficient 
number and location information of the video-level methods.

Although point-level labels contribute to better localizing, their sparse distribution still causes the model to localize inaccurate 
action boundaries or incorrect action snippets \cite{fu2022compact, lee2021learning}. Therefore, several works \cite{ma2020sf, nguyen2019weakly} attempt to directly 
mine snippet-wise pseudo foreground labels by selecting the snippets with high class-specific probabilities, which, however, produces low-quality training pseudo labels. 
The primary issue is that contextual background snippets with high probabilities are often misclassified as foreground labels, resulting in noisy labels \cite{huang2022weakly}.
Additionally, some foreground snippets with low probabilities are wrongly treated as background labels, resulting in fragmentary predictions \cite{lee2021learning}. 

The aforementioned issues motivate us to propose a new solution named multi-refined pseudo label generation (MPLG), aiming to produce more reliable pseudo labels by 
integrating information from class-specific probabilities, attention (class-agnostic) scores, current video features, and point-level labels. To suppress potential 
contextual background snippets with high probabilities while retaining foreground snippets with low probabilities, we use foreground-level feature similarities corresponding 
to point-level labels and current features of the same video to modulate attention scores. Furthermore, we fuse class-level feature similarities corresponding to the same 
categories into class-specific probabilities derived from temporal class activation maps (TCAMs). 

\begin{figure*}[tbp]
  \centering
  \includegraphics[width=\linewidth]{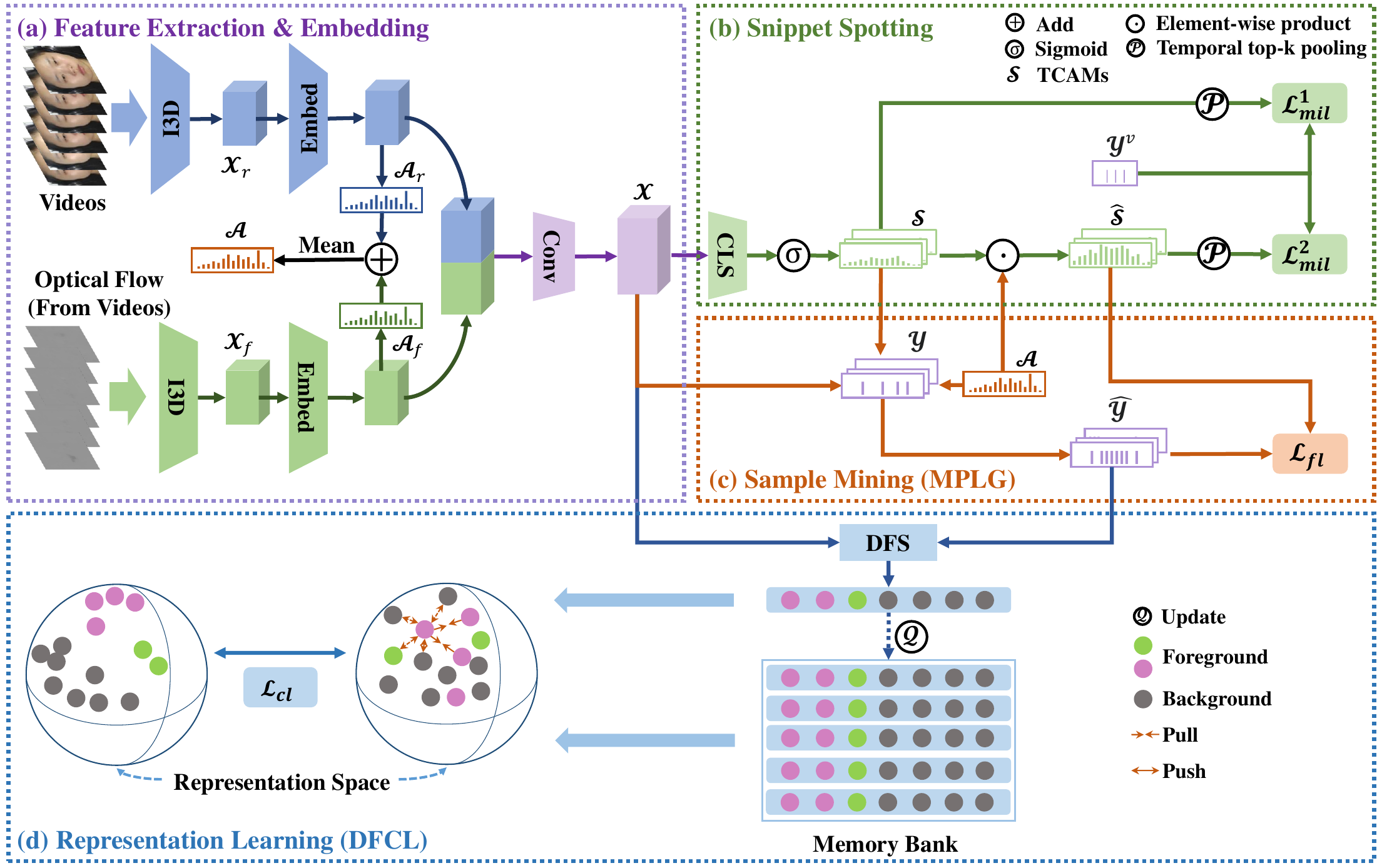}
  \caption{The overall architecture of PWES, which consists of four parts: (a) Feature Extraction and Embedding to utilize a two-stream 
  Inflated 3D ConvNets (I3D) model \cite{carreira2017quo} to exact raw image features $\mathcal{X}_r$ and optical flow features $\mathcal{X}_f$ 
  from uniform non-overlapping snippets. \textcolor{black}{To further extract representational features, we utilize a core saliency compensation module from 
  MC-WES \cite{yu2023weaklysupervised} as our embedding module.} Consequently, these extracted features are individually processed by the embedding module
  and then used to generate 
  attention scores $\mathcal{A}_r$ for raw image modality and attention scores $\mathcal{A}_f$ for optical flow modality, respectively. The 
  mean attention scores $\mathcal{A}$ indicate the probability that the snippet belongs to the foreground. In addition, the processed features 
  are fused as $\mathcal{X}$; (b) Snippet Spotting to process temporal class activation maps (TCAMs) $\mathcal{S}$ and calculate mean 
  class-specific probabilities based on the temporal top-$k$ pooling layer. These mean probabilities are used to compute two multiple instance learning (MIL) losses 
  \cite{Dmaron1997framework}, i.e., $\mathcal{L}_{mil}^1$ and $\mathcal{L}_{mil}^2$ with video-level labels $\mathcal{Y}^v$; (c) Snippet Mining to 
  generate pseudo labels $\mathcal{\widehat{Y}}$ with a multi-refined pseudo label generation (MPLG) algorithm by merging fused video features $\mathcal{X}$, 
  probabilities from TCAMs $\mathcal{S}$, attention scores $\mathcal{A}$, and point-level labels $\mathcal{Y}$. 
  Generated pseudo labels $\mathcal{\widehat{Y}}$ are combined with point-level labels $\mathcal{Y}$ together to calculate the snippet-level classification 
  loss $\mathcal{L}_{fl}$; (d) Representation Learning to implement the distribution-guided 
  feature contrastive learning (DFCL) algorithm with a memory bank. We use a distribution-guided feature sampling (DFS) module to calculate region-level vectors 
  with fused video features $\mathcal{X}$ and pseudo labels $\mathcal{\widehat{Y}}$. Those region-level vectors are used to update the memory bank 
  and calculate the contrastive learning loss $\mathcal{L}_{cl}$.
  }
  \label{fig2}
\end{figure*}  

The feature similarities in MPLG are derived solely from the same videos, without considering the similarities between features corresponding to the same labels across 
the entire dataset. To address this limitation, we propose the distribution-guided feature contrastive learning (DFCL) algorithm, which incorporates a memory bank to 
store enhanced features and facilitates the utilization of contrastive learning to capture global representations. 
The DFCL algorithm guides our spotting model towards convergence in terms of foreground-background separation, inter-class isolation, and intra-class aggregation.

The main contributions of this paper are as follows:
\begin{itemize}
  \item A point-level weakly-supervised expression spotting (PWES) framework is proposed for the first time to achieve frame-level ME and MaE spotting in 
  untrimmed face videos with point-level labels, aiming to alleviate the gap from the time-consuming labeling of frame-level 
  fully-supervised methods and the lack of location and quantitative information of video-level weakly-supervised methods.
  \item For better model training, we propose a MPLG algorithm to generate more reliable pseudo labels by merging class-specific probabilities, attention scores, 
  current video features and point-level labels.  
  \item A DFCL algorithm is introduced to prompt the spotting model to converge towards foreground-background separation, inter-class isolation, 
  and intra-class aggregation by storing video features into the memory bank and implementing feature contrastive learning across the 
  entire datasets with a distribution-guided feature sampling (DFS) module. 
  \item Experimental results on the CAS(ME)$^2$, CAS(ME)$^3$, and SAMM-LV datasets demonstrate that the performance of our weakly-supervised method is quite promising, almost 
  comparable to that of fully-supervised expression spotting models.
  \end{itemize}

The remaining sections are organized as follows. Section \ref{sec2} provides an overview of the related work on expression spotting. In Section \ref{sec3}, 
we present our proposed method in detail. The datasets, implementation details, and experimental results are presented in Section \ref{sec4}. Finally, 
Section \ref{sec5} concludes the paper with future directions.


 

\section{Related Work}
\label{sec2}
\subsection{Fully-supervised Expression Spotting}
Fully-supervised expression spotting methods can be divided as key frame- and interval-based. Key frame-based approaches spot expression intervals by examining 
one or multiple frames in an untrimmed video \cite{pan2020local, zhang2018smeconvnet, yap20213dcnn, liong2021shallow}. In contrast, interval-based methods process 
all the image features from the video as input and focus on the information from neighboring frames. Commonly, long short-term memory (LSTM) is used to encode 
temporal information from neighboring frames in these methods \cite{sun2019two, tran2019dense, verburg2019micro}. However, LSTM may not be effective in handling 
detailed and longer temporal information. To address this issue, several studies \cite{wang2021mesnet, yu2021lssnet, yu2023lgsnet} employ multi-level convolution 
layers or a two-stream network to build long-range temporal dependencies. 

While fully-supervised methods have been successful in expression spotting, they are limited by the requirement of time-comsuming frame-level annotation. In contrast, 
point-level annotations required by our PWES can be obtained with little additional cost, while still achieving frame-level expression spotting. 
Furthermore, our PWES model differs from key frame-based approaches in that it focuses on the simple task of aggregating similar snippets to form expression proposals 
without requiring precise timestamp labels. Key frame-based approaches, however, heavily rely on precise onset and offset annotations to ensure precise localization 
of key-points and boundaries.

\subsection{Weakly-supervised Temporal Action Localization}
Compared with fully-supervised temporal action localization (TAL) methods \cite{lin2017single, yang2020revisiting, zhao2017temporal, zhang2020asfd}, the WTAL methods 
remove the requirement for frame-level annotation and instead utilize video-level \cite{wang2017untrimmednets, nguyen2018weakly, nguyen2019weakly, paul2018w, su2018cascaded} 
or point-level (key frame-level) \cite{lee2021learning, ju2020point, ju2021divide, ma2020sf} labels for training. Video-level WTAL methods can be distinguished 
based on their learning emphasis: foreground-only \cite{wang2017untrimmednets, nguyen2018weakly, paul2018w, su2018cascaded, narayan20193c, hong2021cross}, 
background-assisted \cite{shi2020weakly, liu2019completeness, lee2020background, shi2020weakly, islam2021hybrid}, and pseudo-label-guided \cite{pardo2021refineloc, 
luo2020weakly, he2022asm, nguyen2019weakly, huang2022weakly}. The first type of approaches concentrates on learning the foreground information, while the second 
type enhances background or context learning, and the third type focuses on mitigating the discrepancy between classification and localization. In comparison with 
video-level approaches, point-level WTAL ones \cite{moltisanti2019action, ma2020sf, lee2021learning, ju2021divide} aim to achieve more precise action boundary 
localization by adding a limited amount of supervised information.

Although video-level WTAL methods can significantly reduce the amount of laborious annotations required in fully-supervised settings, they still suffer from the 
absence of information on the location and count of ground truth intervals. As a result, their performances are quite worse compared with the fully-supervised 
approaches. \textcolor{black}{Furthermore, conventional point-level weakly-supervised  methods rely on key-frames, which are still essentially based on the frame-by-frame labeling.
To overcome these limitations, we select a random frame from each ground truth interval into WTAL models as point-level labels, which can supplement insufficient 
information of video-level methods.}

\subsection{Pseudo Label Learning}
The goal of pseudo label learning is to generate ``true'' labels for unlabeled data, which are then combined with ground truth labels to iteratively refine the model during 
training. Typical methods \cite{ma2020sf, lee2021learning, huang2022weakly, pardo2021refineloc, luo2020weakly} regard snippets with high probabilities as pseudo 
training samples, which may inevitably introduce contextual background snippets and overlook foreground snippets. To address this issue, several pseudo label denoising 
methods \cite{lee2021learning, fu2022compact, yang2021uncertainty} utilize probabilities modulated by attention scores or an uncertainty guided collaborative training 
strategy. Nevertheless, these methods still do not leverage attention scores or feature similarities to restrict the range of pseudo label generation or preserve low-confidence 
snippets of the foreground. This is essentially a gap between classification from class-specific probabilities and localization from class-agnostic attention scores. 
Different from the above methods, our MPLG can produce more reliable pseudo labels by merging information from class-specific probabilities, attention scores, current 
video features, and point-level labels.

\subsection{Contrastive Learning}
Contrastive learning \cite{khosla2020supervised, zheng2021weakly, chen2020simple, he2020momentum}, a component of deep metric learning \cite{kaya2019deep}, has demonstrated 
success in various computer vision tasks, such as semantic segmentation \cite{wang2021exploring, zhao2021contrastive} and object detection \cite{tang2021proposal, 
wei2021aligning, xie2021detco}. Its objective is to construct an embedding space where similar instances are grouped together, and different instances are separated using 
noise contrastive estimation (NCE) \cite{gutmann2010noise}. Among various WTAL methods, DCC \cite{li2022exploring} proposes a region-level feature contrast strategy with a region-level 
memory bank to capture global contrast information across the entire dataset. CoLA \cite{zhang2021cola} uses a snippet contrast loss function to refine the hard snippet representation 
in the feature space. To enhance the representation learning of same categories across different videos in expression spotting, our work is the first to introduce the feature 
contrastive learning with a distribution-guided memory bank in our proposed DFCL.

\section{Method}
\label{sec3}
In this section, we present an overview of PWES, as depicted in Figure \ref{fig2}. Specifically, we commence by introducing the problem formulation in Section \ref{sec3.1} and 
outlining the model pipeline in Section \ref{sec3.2}. Subsequently, we delve into the details of MPLG in Sections \ref{sec3.3} and elaborate on DFCL in Section \ref{sec3.4}. 
Furthermore, we provide a comprehensive explanation of the optimization loss terms in Section \ref{sec3.5} and outline the inference process in Section \ref{sec3.6}.

\subsection{Problem Formulation}
\label{sec3.1}
\textcolor{black}{As shown in Figure \ref{fig1}, given an untrimmed video $V=\{{v_\iota }\}_{\iota =1}^L$ has $L$ frames and $N$ point-level annotations with $C$ expression categories. 
Typically, the values of $N$ vary for different training videos. 
During training, the video and its optical flow are first split into non-overlapping uniform snippets of $g$ frames, resulting in $T=L/g$ snippets. The point-level labels are 
represented by $\mathcal{Y} = \{ (\psi_n, y_n)\}_{n=1}^{N}$, where $(\psi_n, y_n)$ represents the labeled timestamp and the corresponding category label for the $n$-th ground truth 
interval in the input video. The label $y_n \in \mathbb{R}^{(C+1)}$ is one-hot vector, and $C+1$ denotes the 
total number of expression categories including the background class. If $y_{n, c}=1$, this means that the $n$-th ground truth interval contains the $c$-th expression instance. 
Furthermore, video-level labels are obtained by aggregating point-level ones along the temporal dimension as $\mathcal{Y}^{v}$, using the indicator function $\mathbbm{1}(\cdot)$, 
where $ y_c^{v} = \mathbbm{1}(\sum_{n=1}^{N} y_{n,c} > 0)$.}

\textcolor{black}{During testing, the spotting model generates $M^{p}$ expression proposals ${E} = \{ \{ f^{on}_{m,c}, f^{off}_{m,c}, y_{m,c}, \phi_{m,c}\}_{m=1}^{M^{p}}\}_{c=1}^{C}$, 
where $f^{on}_{m,c}$ and $f^{off}_{m,c}$ represent the onset and offset frames for the $m$-th expression proposal, respectively, $y_{m,c}$ indicates whether it belongs to the $c$-th 
category, and $\phi_{m,c}$ represents the confidence score belonging to the $c$-th category. As proposals are generated by the class-agnostic attention scores, there are $C$ confidence scores 
for each proposal. Once these proposals are generated, following the existing works \cite{paul2007emotions, wang2021mesnet, yu2021lssnet}, we simplify the task by determining the categories
of the proposals based on their durations and calculating the recall and precision rates, instead of focusing on the classification results.} 

\subsection{Pipeline Overview}
\noindent \textbf{Feature Extraction and Embedding.}
\label{sec3.2}
A two-stream Inflated 3D ConvNets (I3D) model \cite{carreira2017quo} is used to extract image features $\mathcal{X} _{r} \in \mathbb{R} ^{T\times D}$ and optical flow features 
$\mathcal{X}_{f} \in \mathbb{R} ^{T\times D}$ from the divided video snippets. Here, $T$ represents the number of snippets, and $D$ is the feature dimension of a single snippet. 
For the sake of ensuring consistency between the number of raw images and optical flow, the last frame of each video is neglected. Dense optical flow is obtained between adjacent 
frames using the TV-L1 optical flow algorithm \cite{wedel2009improved} with the default smoothing parameter ($\lambda=0.15$). \textcolor{black}{Furthermore, the extracted 
features are processed separately by an embedding module, which is the core saliency compensation module taken from MC-WES \cite{yu2023weaklysupervised}. }
Then the processed features are used to produce the attention scores $\mathcal{A}_r\in \mathbb{R} ^{T}$ for raw image modality and the attention scores $\mathcal{A}_f\in \mathbb{R} ^{T}$ 
for optical flow modality. Consequently, the mean attention scores $\mathcal{A} = 0.5 (\mathcal{A}_r + \mathcal{A}_f)$ are used to indicate the probability that a snippet belongs 
to the foreground. The processed features are also fused and fed into a convolution layer, and the resulting output is denoted as $\mathcal{X}$.

\noindent \textbf{Snippet Spotting.}
The processed features $\mathcal{X}$ are employed in generating temporal class activation maps (TCAMs) $\mathcal{S} \in \mathbb{R} ^{T\times (C+1)}$, where the map $\mathcal{S}$ represents 
the probability distribution of each snippet belonging to all categories via a classifier and a sigmoid function \cite{shou2018autoloc}. Next, we utilize two branches to process 
$\mathcal{S}$ through a temporal top-$k$ pooling layer and obtain the mean class-specific probabilities. One of the two branches is merged with attention scores $\mathcal{A}$ to create 
the suppressed TCAMs $\mathcal{\widehat{S}}$. During testing, these suppressed TCAMs $\mathcal{\widehat{S}}$ and attention scores $\mathcal{A}$ are used to generate expression proposals.

\noindent \textbf{Sample Mining.}
To mine more reliable pseudo training samples, our MPLG algorithm is designed to merge the class-specific probabilities from TCAMs $\mathcal{S}$, 
the attention scores $\mathcal{A}$, the fused features $\mathcal{X}$, and the point-level labels $\mathcal{y}$. In addition, the fused features $\mathcal{X}$ and point-level labels $\mathcal{Y}$ are used 
to calculate foreground-level and class-level feature similarities to re-weight the attention scores $\mathcal{A}$ and the class-specific probabilities.

\noindent \textbf{Representation Learning.}
The purpose of the representation learning in this paper is to capture the global representations across the entire dataset. To this end, we propose a DFCL algorithm based on a 
distribution-guided feature sampling (DFS) module to store region-level features, update region-level features and implement the feature contrastive learning with a memory bank.

\subsection{Multi-refined Pseudo Label Generation}
\label{sec3.3}
Due to the absence of frame-level supervised information and the sparse distribution of point-level labels, existing methods in WTAL \cite{ma2020sf, lee2021learning, fu2022compact} 
often prioritize to mine pseudo labels by selecting snippets with higher probabilities. To reduce more contextual background snippets with high probabilities and 
supplement disregarded snippets with low probabilities in the above pseudo labels, several works \cite{lee2021learning, fu2022compact, yang2021uncertainty} utilize probability 
scores modulated by attention scores or employ an uncertainty guided collaborative training strategy. However, these methods do not fully exploit the localization information to 
effectively narrow down the search space for pseudo foreground labels.

To address this issue, as described in Algorithm \ref{alg1}, we propose a new training sample mining paradigm called multi-refined pseudo label generation (MPLG) algorithm, which 
prioritizes attention scores to reduce unwanted pseudo foreground labels, incorporates feature similarities from point-level labeling counterparts and current video features, and 
ultimately generates pseudo snippet-level labels with class-specific probabilities. Assume the enhanced feature maps from a given video is $\mathcal{X} \in \mathbb{R} ^{T\times D}$ 
with point-level labels $\mathcal{Y} \in \mathbb{R} ^{T\times (C+1)}$. Afterward, the feature maps are fed into a classifier to produce TCAMs $\mathcal{S} \in \mathbb{R} ^{T\times (C+1)}$, 
which are activated with a sigmoid function to generate the snippet-level class-specific probabilities.  
Our MPLG algorithm is elaborated in the following seven steps.
\begin{algorithm*}[htbp]
  \caption{Multi-refined Pseudo Label Generation (MPLG)}
  \label{alg1}
  \LinesNumbered

  \KwIn{enhanced feature maps $\mathcal{X} \in \mathbb{R} ^{T\times D}$, 
  point-level class-specific expression labels $\mathcal{Y} \in \mathbb{R} ^{T\times C}$, 
  snippet-level class-specific scores without background class $\mathcal{\widetilde{S}} \in \mathbb{R} ^{T\times C}$,
  attention scores $\mathcal{A} \in \mathbb{R} ^{T}$, 
  selecting scope $k$,
  predefined threshold $\theta$}
  \KwOut{pseudo labels $\mathcal{\widehat{Y}}$}
  
  \BlankLine
  \tcp {step 1: Merge feature similarities from foreground labels into attention scores; }
  $\mathcal{B}^1 \leftarrow \text{where}(\sum_{i=1}^C \mathcal{Y} _{t,i} > 1) $  \tcp*[l]{Find the indexes corresponding to the point-level labels};
  $\mathcal{G}_{t,n} \leftarrow \mathcal{X}_{t,d} / \| \mathcal{X}_{t} \Vert_2 \otimes \mathcal{X}_{\mathcal{B}^1_d} /\| \mathcal{X}_{\mathcal{B}^1}\Vert_2 $ \; 
  $\mathcal{G}_{t}^{max} \leftarrow \text{argmax}_{i=1}^n(\mathcal{G}_{t,n}) $ \;  
  $\mathcal{G}_{t}^{max} \leftarrow \text{min\_max\_norm}(\mathcal{G}_{t}^{max}) $ \tcp*[l] {the maximum and minimum normalization};
  $\mathcal{A}'_t \leftarrow \mathcal{A}_t \odot \mathcal{G}_{t}^{max} $ \;

  \BlankLine
  \tcp {step 2: Select snippets corresponding to the top-k merged attention scores; }  
  $\mathcal{B}^2 \leftarrow \text{argsort}(\mathcal{A}'_t)[-k:]$ \; 
  $\mathcal{M} \leftarrow [0] * T$  \tcp*[l] {Initialize a vector of length T with all 0s} 
  $\mathcal{M}_{\mathcal{B}^2} \leftarrow  1$\;
  $\mathcal{\widetilde{S}}'_{t,c} \leftarrow \mathcal{\widetilde{S}}_{t,c} \odot \mathcal{M}_t$ \;

  \BlankLine
  \tcp {step 3: Fuse feature similarity from the same categories into TCAMs; } 
  $\mathcal{Z} _{c,d} \leftarrow [0] * C * D$ \;
  \For {$c \in [1, C]$}{
    \If{$\sum_{j=1}^T \mathcal{Y}_{j,c} > 0$}{
      $\mathcal{B}^3 \leftarrow \text{where}( \mathcal{Y}_{j,c} > 1) $ \; 
      $\mathcal{Z} _{c,d} \leftarrow \overline{\mathcal{X}} _{\mathcal{B}^3_d} $ \; 
    }
  }
  $\mathcal{W} _{t,c} \leftarrow \varepsilon(\| \mathcal{X}_{t} -\mathcal{Z} _{c}\Vert_2 )$ \tcp*[l] {If the point-level label for a 
  category does not exist, refer to the discussion in Section \ref{sec3.3}}
  $\mathcal{\widetilde{S}}''_{t,c} \leftarrow \mathcal{\widetilde{S}}'_{t,c} \odot \mathcal{W} _{t,c}$ \;

  \BlankLine
  \tcp {step 4: Delete probabilities less than average one along the temporal dimension; } 
  $\mathcal{\widetilde{S}}''_{t,c}[\mathcal{\widetilde{S}}''_{t,c}< \overline{\mathcal{\widetilde{S}}_{c}}] \leftarrow 0$ \;

  \BlankLine
  \tcp {step 5: Convert refined probabilities to fine-grained pseudo foreground labels; } 
  $\mathcal{\widehat{Y}}^{frg}_{t,c} \leftarrow [0] * T * C$ \;
  $\mathcal{B}^4 \leftarrow \text{where}(\text{argmax}_{i=1}^c(\mathcal{\widetilde{S}}''_{t,c}) > \theta)$ \tcp*[l]{The dimension of $\mathcal{B}^4$ is same as $C$}  
  $\mathcal{\widehat{Y}}^{frg}_{\mathcal{B}^4} \leftarrow 1 $\;

  \BlankLine
  \tcp {step 6: Generate pseudo background labels; } 
  $\mathcal{\widehat{Y}}^{bkg}_{t,1} \leftarrow [0] * T * 1$ \;
  $\mathcal{B}^5 \leftarrow \text{argsort}(-\mathcal{A}_{t})[-2k:]$ \tcp*[l]{The dimension of $\mathcal{B}^5$ is 1}  
  $\mathcal{\widehat{Y}}^{bkg}_{\mathcal{B}^5,1} \leftarrow 1$ \;

  \BlankLine
  \tcp {output: Concatenate pseudo foreground and pseudo background labels} 
  $\mathcal{\widehat{Y}} \leftarrow \text{concatenate}(\mathcal{\widehat{Y}}^{frg}, \mathcal{\widehat{Y}}^{bkg})$
\end{algorithm*}

(1) Merging foreground-level feature similarities corresponding to the point-level labels into the attention scores. We first calculate the similarity matrix between current features 
$\mathcal{X}$ and the selected features $\mathcal{X}_{\mathcal{B}^1}$ \footnote{In this paper, $\mathcal{B}^{\pi}$
represents the index set for the $\pi$-th filtered item.} corresponding to point-level labels $\mathcal{Y}$,
\begin{equation}
  \mathcal{G} =  \frac{\mathcal{X} \otimes \mathcal{X}_{\mathcal{B}^1}}{\| \mathcal{X}\Vert_2 \| \mathcal{X}_{\mathcal{B}^1}\Vert_2},
  \label{equ3.1}
\end{equation}
where $\otimes $ is dot product operator, $\|\cdot \Vert_2$ is a L2-norm function, $\mathcal{B}^1$ means the indexes belonging to the foreground snippets, and 
$\mathcal{G}\in\mathbb{R}^{T\times N}$ is the similarity matrix ($N$ is the number of annotated point-level labels of this video). We select the maximum 
ones $\mathcal{G}^{max} \in \mathbb{R} ^T$ 
from the similarity matrix $\mathcal{G}$ as the modulation factors to merge into the attention scores $\mathcal{A} \in\mathbb{R} ^{T}$ and generate the refined attention scores, 
\begin{equation}
 \mathcal{A}' = \mathcal{A} \odot \mathcal{G}^{max}.
  \label{equ3.2}
\end{equation}
where $\odot$ means element-wise multiplication operator.
These refined attention scores $\mathcal{A}'$ are used to inhibit contextual background snippets in the video. To further amplify the differences between annotated features
and unlabeled features, the above modulation factors are normalized to [0,1]. (The similarity matrix is calculated and filtered 
in a such way that the similarities of the features corresponding to labeled snippets are always 1);

(2) Selecting snippets with the top-$k$ refined attention scores. We select top-$k$ refined attention scores $\mathcal{A}'$ and their indexes $\mathcal{B}^2$ to map onto a mask vector 
$\mathcal{M}_t \in\{0,1\}_{t=1}^T $, where $\mathcal{M}_t=1$ indicates the $t$-th snippet corresponding to a score in top-$k$ attention scores $\mathcal{A}'$. The mask $\mathcal{M}_t$ is used 
as the coarse pseudo foreground labels to exclude a large number of background snippets in TCAMs $\mathcal{\widetilde{S}}$ (snippet-level class-specific scores without background 
class $\mathcal{\widetilde{S}} \in \mathbb{R} ^{T\times C}$) which only includes non-background classes,
\begin{equation}
  \mathcal{\widetilde{S}}'_{t,c}= \mathcal{\widetilde{S}}_{t,c} \odot \mathcal{M}_t,
   \label{equ3.3}
 \end{equation}
where $\mathcal{\widetilde{S}}'$ is the processed TCAMs. 

(3) Fusing class-level feature similarities of the same categories into $\mathcal{\widetilde{S}'}$. After merging the localization information into TCAMs 
$\mathcal{\widetilde{S}}$, which are associated with the classification, we select foreground features corresponding to the same point-level labels with the 
indexes $\mathcal{B}^3$. These selected features are then used to compute mean features along the temporal dimension, 
which serve as the basis for calculating modulation weights to rectify the snippet-level probabilities, 
\begin{equation}
  \mathcal{Z}_c = \overline{\mathcal{X}} _{\mathcal{B}^3},
   \label{equ3.4}
 \end{equation}
\begin{equation}
  \mathcal{W}_{t,c} = \varepsilon (\| \mathcal{X}_{t} -\mathcal{Z}_{c}\Vert_2),
   \label{equ3.5}
 \end{equation}
\begin{equation}
  \mathcal{\widetilde{S}}''_{t,c} = \mathcal{\widetilde{S}}'_{t,c} \odot \mathcal{W}_{t,c},
   \label{equ3.6}
  \end{equation}
where $\mathcal{Z}_c$ represents the mean features of the $c$-th class excluding the background, $\varepsilon$ denotes a softmax function, $\mathcal{W}$ corresponds to 
the modulation weights, and $\mathcal{\widetilde{S}}''$ denotes the modulated TCAMs that incorporate the point-level labeling information.

(4) Deleting snippets with lower probabilities along the temporal dimension. Xu et al. \cite{xu2016heterogeneous} observe that the majority of frames belong to 
the background. Therefore, we proceed to exclude multi-refined snippets that have high probabilities belonging to the background. To simplify the process, we 
discard the snippets with lower probabilities from $\mathcal{\widetilde{S}}''$,
\begin{equation}
  \mathcal{\widetilde{S}}''_{t,c}[\mathcal{\widetilde{S}}''_{t,c}< \overline{\mathcal{\widetilde{S}}_{c}}] = 0,
   \label{equ3.7}
  \end{equation}
where $\overline{\mathcal{\widetilde{S}}_{c}}  = \frac{1}{T} \sum_{t = 1}^{T} \mathcal{\widetilde{S}}_{t,c}$ represents the average probability scores from the 
original TCAMs $\mathcal{\widetilde{S}}$.

(5) Converting refined class-specific scores to fine-grained pseudo foreground labels. Consistent with previous works \cite{ma2020sf, li2022exploring, 
fu2022compact}, a snippet in video $\mathcal{V}$ is classified to class $c'$ and included in the training set if the class-specific probability 
score from TCAMs $\mathcal{\widetilde{S}}''_{c'}$ exceeds a predefined threshold $\theta$. Mathematically, the pseudo foreground label for the 
$j$-th snippet can be calculated by,
\begin{equation}
  \mathcal{\widehat{Y}}^{frg}_c =
  \begin{cases}
    1, & if \ c== \mathop{argmax}\limits_{c',j} \mathcal{\widetilde{S}}''_{c',j} \& \ \mathcal{\widetilde{S}}''_{c',j} > \theta \\
    0, & otherwise. 
    \end{cases}
  \label{equ3.8}
\end{equation}

(6) Generating pseudo background labels. Considering the sparsity of the foreground in video $\mathcal{V}$, we select the number of the
negative training samples to be twice that of positive training samples in the original attention scores $\mathcal{A}$. The pseudo background 
labels are generated as,
\begin{equation}
  \mathcal{\widehat{Y}}^{bkg}_t =
  \begin{cases}
    1, & if \ t \in argsort(-\mathcal{A}_{t})[2k:] \\
    0, & otherwise.
    \end{cases}
  \label{equ3.9}
\end{equation}

(7) Outputting all pseudo labels. The pseudo foreground labels $\mathcal{\widehat{Y}}^{frg}$ and pseudo background labels $\mathcal{\widehat{Y}}^{bkg}$ are 
concatenated along the class dimension to obtain the output $\mathcal{\widehat{Y}}$, which is then used to calculate the snippet-level classification loss in 
conjunction with the re-modulated TCAMs $\mathcal{\widehat{S}}$.

Since some foreground snippets may have low probabilities, we regard them as ambiguous samples. Inspired by the online hard example mining (OHEM) algorithm 
\cite{Shrivastava2016training} in object detection \cite{Girshick2012fast}, we adopt a similar approach to handle these ambiguous samples by
limiting the inclusion of snippets used for generating pseudo labels. 
Specifically, we ensure that the value of $k$ (the number of selected snippets used for generate coarse pseudo foreground labels) 
is no more than 0.3 times of the total number of snippets in the current video $\mathcal{V}$. Additionally, if a point-level label for a particular category 
is absent in video $\mathcal{V}$, we assign a value of 0 to $\mathcal{W}$ for the current category, while setting it as 1.0 for the other categories.

\subsection{Distribution-guided Feature Contrastive Learning}
\label{sec3.4}
Existing expression spotting methods often focus on video features within specific video, neglecting the global features of the same labels across multiple videos 
of the entire dataset \cite{verburg2019micro, wang2021mesnet, yu2021lssnet}. In contrast, we introduce a feature contrastive learning strategy \cite{khosla2020supervised, 
zheng2021weakly} relying on a memory bank to capture global representations. However, directly applying video features of varying durations for contrastive learning 
may significantly complicate model optimization. Hence, we address this challenge by extracting fixed-length features from each video to store in the memory bank.
A straightforward approach is to preprocess the input video features by sampling them to a fixed size. However, such sampling may alter the duration of ground 
truth intervals, causing the frames corresponding to the point-level labels to lose their representational cues. This, in turn, will negatively impact the subsequent 
predictions, given that the final classification is based on proposal duration \cite{paul2007emotions}. In fact, the point-level WTAL methods also follow 
a no-sampling approach in training videos \cite{fu2022compact, lee2021learning}.

\textcolor{black}{Considering these factors, we propose to sample the enhanced features processed by our model during training. DCC \cite{li2022exploring} suggests that video features of 
the same size can be divided into evenly-sized region-level features to accelerate the generation of global representations, where each mean region-level vector is calculated 
along the temporal dimension for feature reduction. This inspired us to adopt a similar way, i.e., regardless of the video durations, we only store same number of 
region-level vectors from each video into the memory bank. In terms of the number of vectors preserved in each category, we mainly rely on the prior distribution of categories.
To achieve this, we design the DFCL algorithm, which constructs a memory bank to store region-level vectors from all videos in the dataset and captures global representational 
features through contrastive learning.}
Specifically, we generate three video feature sets by element-wise filtering of the fused video features $\mathcal{X}$ with pseudo labels $\mathcal{\widehat{Y}}$,
\begin{equation}
  \mathcal{\widehat{X}}_c = \mathcal{X} \odot \mathcal{\widehat{Y}}_c.
   \label{equ3.10}
\end{equation}
\textcolor{black}{Since the distributions of MEs, MaEs and background are different, we design a distribution-guided feature sampling (DFS) module to further divide each of three 
feature sets into a different number of region-level spaces, where the size of region-level spaces is same for each category. The number of spaces is determined based on the distribution 
information of categories, with a minimum number allocated to the ME feature set.} This is because MEs are more susceptible to harsh excitation conditions compared to MaEs \cite{qu2017cas, 
yap2020samm}. Conversely, the background feature set is assigned with the maximum number of spaces, considering its dense presence \cite{xu2016heterogeneous}. Therefore, the region-level 
spaces for the $c$-th class are defined as,
\begin{equation}
  \mathcal{\widehat{X}}_{c,k} \Rightarrow \{\mathcal{Q}_{c,k} \}_{k=1}^{K_c},
   \label{equ3.11}
\end{equation}
where $K_c$ is the number of region-level spaces for the $c$-th class. To preserve the global representations of individual video features, we treat all feature set
$\mathcal{\widehat{X}}_{c}$ for the $c$-th class as a main region-level space $\mathcal{Q}_{c,0}$. We then temporally pool region-level spaces by averaging 
to form region-level vectors, i.e., $q_{c,0}, \cdots, q_{c,K_c}$. 
Concatenating all region-level vectors $q = [q_{0,0}, \cdots, q_{0, K_0}, \cdots , q_{C+1, 0},\cdots, q_{C+1, K_{C+1}}]$ as fixed-size feature maps extracted from 
each video, which are used to update in the memory bank and implement feature contrastive learning for learning global representations across the entire dataset. The 
DFCL algorithm ultimately enables our spotting model to generate reliable pseudo labels and converge towards foreground-background separation, inter-class isolation, 
and intra-class aggregation.

\subsection{Optimization Loss}
\label{sec3.5}
\noindent \textbf{Video-level Classification Loss.}
As shown in Figure \ref{fig2}(b), there are two branches to process the probabilities from TCAMs $\mathcal{S}$. The first branch is based on original probabilities,  
using a temporal top-k pooling layer and video-level distribution consistency strategy \cite{yu2023weaklysupervised} to produce video-level class confidence scores $p_i^1$ 
for the $i$-th category. We calculate the multiple instance learning (MIL) loss with the video-level labels $\mathcal{Y}^{v}$ in this branch as:
\begin{equation}
  \mathcal{L}_{mil}^1 = -\sum_{c=1}^{C+1} \mathcal{Y}^{v} \log (p_i^1).
  \label{equ3.12}
\end{equation}
The second branch applies the attention scores $\mathcal{A}$ to inhibit the background snippets in TCAMs $\mathcal{S}$ to generate video-level class confidence scores $p_i^2$ 
with the processed TCAMs $\mathcal{\widehat{S}}$. The MIL loss of this branch using video-level labels $\mathcal{\widehat{Y}}^{v}$ without the background class is denoted as:
\begin{equation}
  \mathcal{L}_{mil}^2 = -\sum_{c=1}^{C} \mathcal{\widehat{Y}}^{v} \log (p_i^2).
  \label{equ3.13}
\end{equation}
Therefore, the video-level classification loss is defined as:
\begin{equation}
  \mathcal{L}_{vcl} = \mathcal{L}_{mil}^1 + \mathcal{L}_{mil}^2.
  \label{equ3.14}
\end{equation}

\noindent \textbf{Snippet-level Classification Loss.}
Once pseudo labels $\mathcal{\widehat{Y}}$ are obtained, we can calculate the snippet-level classification loss by combining the point-level labels 
$\mathcal{Y}$ together to learn fine-grained localization information. In details, the snippet-level labels combined using the indicator function are denoted as:
\begin{equation}
  \mathcal{\widetilde{Y}}_{t,c} = \mathbbm{1}((\mathcal{Y}_{t,c} + \mathcal{\widehat{Y}}_{t,c}) > 0),
  \label{equ3.15}
\end{equation}
where $\mathcal{\widetilde{Y}}$ is the snippet-level combined labels.
We then constitute snippet-level probabilities as $\mathcal{\hat{S}}$ by concatenating the foreground probabilities from processed TCAMs $\mathcal{\widehat{S}}$ 
and the background probabilities from TCAMs $\mathcal{S}$. Therefore, the snippet-level classification loss is defined as:
\begin{equation}
  \begin{split}
  \mathcal{L}_{scl} & = -\frac{1}{N_s} \sum_{t=1}^{N_s}\sum_{c=1}^{C+1} (\mathcal{\widetilde{Y}}_{t,c} (1-\mathcal{\hat{S}}_{t,c})^2 \log \mathcal{\hat{S}}_{t,c} \\
  & + (1-\mathcal{\widetilde{Y}}_{t,c}) {\mathcal{\hat{S}}}_{t,c}^2 \log (1-\mathcal{\hat{S}}_{t,c})),
  \label{equ3.16}
  \end{split}
\end{equation}
where $N_s$ is the number of snippets containing snippet-level labels in $\mathcal{\widetilde{Y}}$.

\noindent \textbf{Feature Contrastive Learning Loss.}
As described in Section \ref{sec3.4}, given the region-level vectors $q$ from one video, we can calculate the feature contrastive learning loss based on the 
memory bank where we construct positive and negative feature sets across all videos. 
In details, given a region-level vector region as $u_h$, the positive feature set $\mathcal{P}^{+}$ for $u_h$ contains region-level vectors corresponding 
to the same foreground labels, while the negative feature set $\mathcal{N}^{-}$ for $u_h$ contains ones corresponding to the different foreground 
labels and the background. For simplicity, the region-level vectors $q$ is normalized by a L2-Norm function as $\widehat{q}$. Following the InfoNCE loss 
\cite{he2020momentum}, our feature contrastive learning loss is defined as
\begin{equation}
  \mathcal{L}_{fcl} = -\frac{1}{N_c}\sum_{h=1}^{N_c}\log(\frac{\sum_{\widehat{q}_h^{+} \in \mathcal{P}^{+}}\exp(u_h^{} \cdot \widehat{q}_h^{+}/\tau)}{
    \sum_{\widehat{q}_h^{} \in {\mathcal{P}^{+} \bigcup \mathcal{N}^{-}}}\exp(u_h^{}  \cdot \widehat{q}_h^{} /\tau)} ),
  \label{equ3.17}
\end{equation}
where $N_c$ represents the number of samples in positive set $\mathcal{P}^{+}$, and $\tau$ is a temperature scale.

\noindent \textbf{Final Joint Loss.}
Our model incorporates the guide loss function $\mathcal{L}{gui}$ and the sparsity loss function $\mathcal{L}{sps}$, which have been utilized in previous 
works \cite{islam2021hybrid, hong2021cross, yu2023weaklysupervised}. The former ensures the mutual exclusiveness between attention scores and the probabilities 
of snippets belonging to the background. The latter promotes sparsity in attention scores. Furthermore, we employ the mutual learning loss $\mathcal{L}_{aml}$, 
introduced in CO$_2$-Net \cite{hong2021cross}, to modality-specific attention scores using the mean square error (MSE) function. 
All the above mentioned loss functions are combined to form the final optimization function,

\begin{equation}
  \begin{split}
    \mathcal{L}_{all} & =  \lambda_1\mathcal{L}_{vcl} + \lambda_2\mathcal{L}_{scl} + \lambda_3\mathcal{L}_{fcl} \\
    & + \lambda_4\mathcal{L}_{gui} + \lambda_5\mathcal{L}_{sps} + \mathcal{L}_{aml},
  \end{split}
  \label{equ3.18}
\end{equation}
where $\lambda_1$, $\lambda_2$, $\lambda_3$, $\lambda_4$, and $\lambda_5$ are hyperparameters.

\subsection{Inference Process}
\label{sec3.6}
During testing, in line with MC-WES \cite{yu2023weaklysupervised}, we employ a multi-top method, constructing a series of consecutive integers. Each value in this series 
is utilized to select snippets along the temporal dimension. These selected snippets are then sorted based on timestamps. Consequently, the snippets corresponding to 
consecutive timestamps form the final proposals. Regarding confidence scores, we employ the approach presented in AutoLoc \cite{shou2018autoloc}, computing outer-inner 
scores belonging to different categories as the ultimate confidence scores.

\section{Experiments}
\label{sec4}
\subsection{Datasets}
\label{sec4.1}
We evaluate our framework on three popular datasets for facial expression spotting: CAS(ME)$^2$ \cite{qu2017cas}, CAS(ME)$^3$ \cite{li2022cas}, and SAMM-LV \cite{yap2020samm}. 
CAS(ME)$^2$ consists of 98 long videos with an average length of 2940 frames captured at 30 FPS, with 96\% of the frames being the background. This dataset is annotated 
with 57 ME and 300 MaE intervals. SAMM-LV contains of 224 long videos with an average length of 7000 frames captured at 200 FPS, with 68\% of 
the frames being the background. This dataset is annotated with 159 MEs and 340 MaEs. The CAS(ME)$^3$ dataset consists of 956 videos with an 
average length of 2600 frames at 30 FPS, with 84\% of the frames being the background. This dataset is annotated with 207 MEs and 2071 MaEs.

\subsection{Evaluation Metric}
\label{sec4.2}
Following Micro-Expression Grand Challenge (MEGC) 2019 \cite{see2019megc} and 2022 \cite{li2022megc2022}, a predicted expression proposal $E$ is considered as a true 
positive (TP) sample only if there exists a ground truth interval $E_{gt}$ that satisfies
\begin{equation} 
  \frac{E\bigcap E_{gt}}{E\bigcup  E_{gt}} \geq k_{eval},
  \label{equ3.19}
\end{equation} 
where $k_{eval}$ is the intersection over union (IoU) threshold, which is commonly predefined set to 0.5 for expression spotting. 

According to the MC-WES evaluation protocol \cite{yu2023weaklysupervised}, F1-scores are calculated for all proposals, as well as for MEs using three different 
versions: F1(0.5), F1(1.0), and F1(p). F1(0.5) measures the ME proposals with durations shorter than 0.5 seconds among all expression proposals. 
F1(1.0) evaluates the ME proposals lasting less than 1.0 second among all expression proposals. Additionally, F1(P) assesses the ME proposals shorter 
than 0.5 seconds, corresponding to the proposal set with the optimal overall F1-score.

\subsection{Implementation Details}
\noindent \textbf{Input Feature Extraction.}
For the CAS(ME)$^2$, CAS(ME)$^3$, and SAMM-LV datasets, we first split videos and optical flow into non-overlapping snippets. Specifically, 
we sample continuous non-overlapping 8 frames as a snippet for the CAS(ME)$^2$ and CAS(ME)$^3$ datasets, and 32 frames as a snippet for the SAMM-LV dataset \cite{yu2023weaklysupervised}.
Then, the I3D model \cite{carreira2017quo} is applied to extract 1024-dimension features for each snippet. 

\noindent \textbf{Training Details.}
We train our PWES using the common strategy of leave-one-subject-out (LOSO). The original number of snippets in each video is used as $T$ without any additional sampling. During training, 
we use a batch size of 8 and employ the Adam optimizer \cite{kingma2014adam}. The model is trained for 1000 iterations for each dataset. We set the following values for the 
hyperparameters: $\lambda_1=\lambda_2 =\lambda_4=1.0, \lambda_3 = 0.1, \lambda_5=0.8$, and $k=0.3$ for the three datasets. 

Our approach involves several steps, starting with pretraining 
iterations to train the model. Subsequently, we execute the MPLG and DFCL algorithms to generate reliable pseudo labels and construct a  memory bank.
For the CAS(ME)$^2$ and CAS(ME)$^3$ dataset, we perform 80 pretraining iterations and set the learning rate to 0.0005. In the MPLG and DFCL algorithms, we use $\theta=0.8$, $\tau=1.0$, 
and include 10 region-level background vectors, 2 region-level ME vectors and 3 region-level MaE vectors.
For the SAMM-LV dataset, we conduct 50 pretraining iterations with a learning rate of 0.0008. In the MPLG and DFCL algorithms, we set $\theta=0.9$, $\tau=1.0$, and include 8 region-level 
background vectors, 2 region-level ME vectors and 3 region-level MaE vectors.

\noindent \textbf{Testing Details.}
We use non-maximum suppression (NMS) \cite{neubeck2006efficient} to remove redundant proposals with a threshold of 0.01 on the three datasets. The remaining proposals 
are used to calculate recall rate, precision rate, and four versions of F1-scores.

\subsection{Ablation Study and Analysis}
In this section, we conduct extensive experiments on CAS(ME)$^2$ for ablation study and analysis of our method.

\subsubsection{Effectiveness of Main Components}
MPLG and DFCL are two main components in our proposed PWES framework shown in Figure \ref{fig2}. Specifically, Our MPLG is designed to generate more reliable pseudo labels and alleviate 
the sparsity of point-level labels in each video, and DFCL is designed to capture the global feature representation across all dataset. Detailed 
results are shown in Table \ref{tab1}. The ``Baseline'' refers to the baseline model derived from MC-WES \cite{yu2023weaklysupervised} without the attention-guided feature consistency loss 
and attention-guided duration consistency loss. The ``PPLM'' represents the results of our reproduction based on CRRC-Net \cite{fu2022compact}, as no released code is available for it.

As shown in Table \ref{tab1}, when MPLG is added into the baseline model, the performance has a more significant improvement than that of the model with the addition of the PPLM module \cite{fu2022compact}, 
with a 6.8\%, 8.9\%, and 7.8\% improvement in recall, precision, and F1-scores for all proposals, respectively. Introducing MPLG and DFCL to the baseline model further increases the recall 
rate and F1-score for all proposals by 3.3\% and 2.1\%, respectively, albeit with a slight reduction in precision rate (from 0.473 to 0.476). This can be attributed to MPLG's ability to substantially reduce unreliable 
pseudo labels, while having limited capacity to spot more reliable proposals. DFCL can compensate for this by capturing global representations without introducing excessive noise, thereby 
improving the recall of all proposals.

Regarding ME spotting, our MPLG notably improves the localization capability, although it does not significantly impact the overall optimal 
proposal set. Conversely, the introduction of both components further enhances ME spotting in the overall optimal proposal set, as illustrated by the values of F1(p).

\begin{table}[htbp]
  \centering
  \normalsize
  \setlength\tabcolsep{3pt}
  \caption{Performances with different components on the CAS(ME)$^2$ dataset. ``REC'' is the recall rate, ``PRE'' is the precision rate, and ``F1'' is 
            the F1-score.}
    \begin{tabular}{l|cccccc}
    \toprule
    Algorithm                 &F1(0.5)&F1(1.0)&F1(p)&REC&PRE&F1 \\
    \midrule
    Baseline                  &0.167 &0.076 &0.000 &0.252 &0.282 &0.266 \\
    +PPLM\cite{fu2022compact} &0.172 &0.081 &0.000 &0.204 &0.384 &0.267 \\
    +MPLG                     &\textbf{0.207} &0.095 &0.000 &0.272 &\textbf{0.473} &0.345 \\
    +MPLG+DFCL                &0.203 &\textbf{0.102} &\textbf{0.023} &\textbf{0.305} &0.467 &\textbf{0.366} \\
    \bottomrule
    \end{tabular}
  \label{tab1}
\end{table}

\subsubsection{Compositions in MPLG}
To provide a more detailed understanding of the multi-refinement differences in our MPLG designed for generating pseudo labels, we compare its performance with different pseudo label 
generation strategies, as presented in Table \ref{tab2}. The ``MPLG-c'' strategy exclusively relies on class-specific probabilities from TCAMs to generate pseudo labels. The 
results indicate that this strategy does not improve or even slightly decreases the model's overall performance, especially the recall decreased by 2.5\%, resulting in a 0.6\% decline 
in the F1-score. One reason for this is the notable disparity between the attention scores related to localization and the probability scores related to classification. Furthermore, we 
observe that the positive sample distribution is sparse, indicating a scarcity of high localization-related attention scores. Consequently, the ``MPLG-ca'' strategy is tested, 
which incorporates both class-specific probabilities and attention scores to generate pseudo labels. 
The results show that this strategy can improve the precision by 1.7\% and eventually improve overall F1-score to 0.27, while maintaining a slight decrease in recall 
compared to the ``MPLG-c'' strategy and the baseline. Nevertheless, directly filtering attention scores still fails to bridge the gap between localization-related attention scores and 
classification-related probabilities, particularly in terms of selecting high attention scores corresponding to contextual backgrounds.

Considering the utilization of point-level labels for training our model, the current video features associated with these labels serve as a guide to emphasize similarities within the 
same classes and differences across different classes. To accomplish this, we design the ``MPLG-caf1'' strategy to integrate foreground-level similarities, derived from point-level labels 
and current video features, into the attention scores. We then select the top-$k$ enhanced attention scores to generate pseudo foreground labels in combination with class-specific probabilities. 
As shown in Table \ref{tab2}, the ``MPLG-caf1'' strategy demonstrates a significant improvement in recall, precision, and overall F1-score, achieving 7.3\%, 6.7\%, and 7.0\% respective increases compared to the ``MPLG-ca'' 
strategy. The results effectively validate the effectiveness of our current strategy. Similarly, we construct the ``MPLG-caf2'' strategy, where class-level similarities are computed from 
point-level labels and current video features, and then integrated with class-specific probabilities to generate pseudo foreground labels. The results demonstrate that the ``MPLG-caf2'' strategy 
can further improve the precision from 0.365 to 0.416 compared to the ``MPLG-caf1'' strategy, but reduce the recall by 10.3\%, which also leads to a lower F1-score.

Finally, we fuse the ``MPLG-caf1'' and ``MPLG-caf2'' strategies to implement our MPLG algorithm. The outcomes demonstrate that the precision can be further improved to 0.473 in comparison to 
the ``MPLG-caf1'' strategy, but concurrently, there is a substantial decrease in recall. From the above 
analysis, we can conclude that integrating information from class-specific probabilities, attention scores, current video features, and point-level labels is essential for generating reliable 
pseudo labels that can effectively enhance the overall localization capability of the model.

The results of ME spotting corresponding to the MPLG algorithm are consistent with the overall outcomes, except for the F1(p) scores. The main reason for this discrepancy is that 
incorporating class-level similarities into class-specific probabilities can hinder the ME spotting capability of our model, as evidenced by the results obtained from the ``MPLG-caf2'' strategy. 
Essentially, MaEs tend to have a denser distribution compared to MEs in most videos \cite{yan2014casme}. Consequently, the bias towards class-level similarities favors MaEs and 
suppresses class-specific probabilities that should correspond to MEs in TCAMs.

\begin{table}[htbp]
  \centering
  \normalsize
  \setlength\tabcolsep{3pt}
  \caption{Performances with different pseudo label generation strategies on the CAS(ME)$^2$ dataset.}
    \begin{tabular}{l|cccccc}
    \toprule
    Algorithm    &F1(0.5)&F1(1.0)&F1(p)&REC&PRE&F1 \\
    \midrule
    Baseline     &0.167 &0.076 &0.000 &0.252 &0.282 &0.266 \\
    MPLG-c       &0.206 &0.081 &0.000 &0.227 &0.305 &0.260 \\
    MPLG-ca      &0.172 &0.073 &0.000 &0.246 &0.298 &0.270 \\
    MPLG-caf1    &0.200 &0.088 &\textbf{0.087} &\textbf{0.319} &0.365 &0.340    \\
    MPLG-caf2    &0.172 &0.080 &0.000 &0.216 &0.416 &0.284 \\
    MPLG         &\textbf{0.207} &\textbf{0.095} &0.000 &0.272 &\textbf{0.473} &\textbf{0.345} \\
    \bottomrule
    \end{tabular}
  \label{tab2}
\end{table}

\subsubsection{Hyperparameters in MPLG}
To generate more reliable pseudo labels, we first pretrain our model with a specified number of iteration and then execute the MPLG algorithm. We vary the 
number of pretraining iteration in Table \ref{tab3} to analyze its effect on the model's performance. Additionally, to evaluate the impact of the predefined 
threshold $\theta$ on generating foreground samples, we experiment with different $\theta$ values in Table \ref{tab4}. Finally, we examine the effect of using 
different proportions of snippets to generate pseudo labels on model training by testing various $k$ values in Table \ref{tab5}.

\noindent \textbf{Pretraining iteration.}
Table \ref{tab3} indicates that after 80 pretraining iterations, our model achieves optimal recall, accuracy, and F1-score at an overall level, demonstrating its robust 
capability (optimal F1 (p)) in global ME spotting within the optimal proposal set. However, it does not yield the best results in terms of ME spotting, specifically in F1(0.5) 
and F1(1.0). Furthermore, conducting too many iterations significantly compromises the reliability of the generated pseudo labels, although it enhances ME spotting. One 
potential explanation for this phenomenon is the scarcity of training data for MEs in comparison to the training data available for MaEs. Excessive iterations tend to bias 
the model's spotting towards MaEs.

\begin{table}[htbp]
  \centering
  \normalsize
  \setlength\tabcolsep{4pt}
  \caption{Performances with different number of pretraining iteration on the CAS(ME)$^2$ dataset.}
    \begin{tabular}{l|cccccc}
    \toprule
    Iteration    &F1(0.5)&F1(1.0)&F1(p)&REC&PRE&F1 \\
    \midrule
    30     &0.203  &0.093  &0.000  &0.266  &0.459  &0.337 \\
    50     &0.195  &0.080  &0.050  &0.275  &0.455  &0.342 \\
    80     &0.203  &0.102  &\textbf{0.023} &\textbf{0.305} &\textbf{0.467} &\textbf{0.366} \\
    100    &\textbf{0.230}  &0.103  &0.000  &0.289  &0.428  &0.343 \\
    120    &0.196  &\textbf{0.107} &0.000  &0.276  &0.451  &0.342 \\
    \bottomrule
    \end{tabular}
  \label{tab3}
\end{table}

\noindent \textbf{Predefined threshold.}
As shown in Table \ref{tab4}, the role of $\theta$ within MPLG is evident. 
Lower threshold values tend to introduce the omission of numerous foreground snippets, leading to shorter proposal intervals, which can enhance ME spotting but may reduce 
overall recall and F1-score.
Conversely, higher threshold values prioritize the localization of a greater number of foreground snippets, leading to better overall recall, precision 
and F1-score but potentially sacrificing ME spotting capability. Within the range of 0.8 to 0.9 for $\theta$, the model demonstrates stable performance.
Notably, the best spotting performance is achieved when $\theta$ is set to 0.8, suggesting the optimal balance between ME and overall performance.

\begin{table}[htbp]
  \centering
  \normalsize
  \setlength\tabcolsep{5pt}
  \caption{Performances with different predefined threshold $\theta$ for MPLG on the CAS(ME)$^2$ dataset. }
    \begin{tabular}{l|cccccc}
    \toprule
    $\theta$    &F1(0.5)&F1(1.0)&F1(p)&REC&PRE&F1 \\
    \midrule
    0.5    &0.229  &0.093  &0.000  &0.274  &0.379  &0.343\\
    0.6    &\textbf{0.233} &\textbf{0.114}  &0.000  &0.277  &0.460  &0.352 \\
    0.7    &0.219  &0.095  &0.000  &0.283  &0.409  &0.351 \\
    0.8    &0.203  &0.102  &\textbf{0.023} &\textbf{0.305} &\textbf{0.467} &\textbf{0.366} \\ 
    0.9    &0.226  &0.084  &0.000  &0.302  &0.460  &0.360 \\
    \bottomrule
    \end{tabular}
  \label{tab4}
\end{table}

\noindent \textbf{Top-$k$ value.}
Considering the presence of ambiguous samples, which are more challenging to learn, we employ an approach inspired by the OHEM algorithm \cite{Shrivastava2016training} and refrain from 
labeling all snippets in MPLG of Section \ref{sec3.3} (2). \textcolor{black}{Considering the number of negative snippets is much larger than that of positive snippets, in order to maintain
sufficient positive snippets in generating pseudo labels, this paper empirically sets the number of negative samples in all experiments to be 
twice the number of positive samples. Specifically, we choose the snippets corresponding to the top 30\% (e.g., $k=0.3$) of the attention scores as positive samples, and the snippets
corresponding to the last 60\% of the attention scores as negative samples in the current video.} 
The results presented in Table \ref{tab5} highlight the significance of the value of $k$ in achieving desirable outcomes. A smaller $k$ (e.g., 0.1 or 0.2) reduces overall spotting performance 
but achieves the optimal F1-score for MEs. \textcolor{black}{This selection of a smaller $k$ indicates that only the snippets with larger attention scores are selected, while those key 
snippets with lower attention scores may be excluded.} This results in fragmentary and inaccurate predictions, leading to lower overall recall, precision, and F1-score. 
Conversely, a larger $k$ yields better overall spotting results, particularly for recall and F1-score. However, a larger $k$ introduces more noisy ME snippets and reduces
the recall, precision, and F1-score of MEs. While a compromised $k$ leads to the best overall precision, it significantly degrades the performance of MEs and fails to achieve optimal 
recall and F1-score of MEs. Thus, we can draw a conclusion that expanding the search range can capture some MaE snippets with low attention scores. However, for ME snippets, 
even with an expanded search range, our method is unable to identify snippets with low attention scores. Additionally, when we choose larger $k$ to expand search range, too many 
noisy snippets may be introduced, which will diminish the model's spotting capability.

\begin{table}[htbp]
  \centering
  \normalsize
  \setlength\tabcolsep{5pt}
  \caption{Performances with different top-$k$ values on the CAS(ME)$^2$ dataset.}
    \begin{tabular}{l|cccccc}
    \toprule
    $k$    &F1(0.5)&F1(1.0)&F1(p)&REC&PRE&F1 \\
    \midrule
    0.1     &\textbf{0.233} &\textbf{0.113}  &\textbf{0.034}  &0.246  &0.456  &0.320 \\
    0.2     &0.200  &0.089  &0.000  &0.246  &\textbf{0.524}  &0.335 \\
    0.3     &0.203  &0.102  &0.023 &\textbf{0.305} &0.467 &\textbf{0.366} \\
    \bottomrule
    \end{tabular}
  \label{tab5}
\end{table}

\subsubsection{Hyperparameters in DFCL}
To analyze the performance of DFCL in capturing global representations, we test to vary the value of temperature scale $\tau$, as presented 
in Table \ref{tab6}. Additionally, we assess the impact of the number of background region-level vectors on feature contrastive learning, as 
shown in Table \ref{tab7}.

\noindent \textbf{Temperature scale $\tau$.}
The temperature scale $\tau$ plays a crucial role in regulating attention allocation to challenging samples, with smaller values emphasizing the distinction 
between such samples and their most similar counterparts \cite{wang2021understanding}. As shown in Table \ref{tab6}, smaller values of $\tau$ result in poorer 
performance for both ME spotting and overall expression spotting. Conversely, larger values of $\tau$ lead to better performance, particularly demonstrating 
a notable positive correlation with overall recall. Nevertheless, excessively large $\tau$ negatively affect the model's precision and F1-score. 
Therefore, we conclude that the difference in features processed by our model is significant between MEs and MaEs. This suggests 
that there is no need to use too small temperature scale when learning fine-grained features.

\begin{table}[htbp]
  \centering
  \normalsize
  \setlength\tabcolsep{5pt}
  \caption{Performances with different temperature scale $\tau$ on the CAS(ME)$^2$ dataset.}
    \begin{tabular}{l|cccccc}
    \toprule
    $\tau$  &F1(0.5)&F1(1.0)&F1(p)&REC&PRE&F1 \\
    \midrule
    0.07    &0.174  &0.078  &0.000  &0.277  &0.429  &0.337 \\
    0.2     &0.203  &0.095  &0.000  &0.272  &0.462  &0.342 \\
    0.5     &0.230  &0.096  &0.000  &0.277  &0.458  &0.346 \\
    0.8     &\textbf{0.241}  &0.092  &0.000  &0.303  &0.417  &0.351 \\
    1.0     &0.202  &\textbf{0.102}  &\textbf{0.023}  &0.306  &\textbf{0.467}  &\textbf{0.366} \\
    1.5     &0.185  &0.079  &0.000  &\textbf{0.336}  &0.450  &0.343 \\
    \bottomrule
    \end{tabular}
  \label{tab6}
\end{table}

\begin{table}[htbp]
  \centering
  \normalsize
  \setlength\tabcolsep{5pt}
  \caption{Performances with different number of background region-level vectors on the CAS(ME)$^2$ dataset. }
    \begin{tabular}{l|cccccc}
    \toprule
    Number   &F1(0.5)&F1(1.0)&F1(p)&REC&PRE&F1 \\
    \midrule
    5     &0.182  &0.100  &0.000  &0.289  &0.410  &0.339 \\
    8     &0.203  &0.088  &0.000  &0.303  &0.441  &0.359 \\
    10    &0.202  &0.102  &\textbf{0.023}  &0.305  &\textbf{0.467} &\textbf{0.366} \\
    12    &0.200  &\textbf{0.129}  &0.000  &\textbf{0.313}  &0.401  &0.352 \\
    15    &\textbf{0.226}  &0.095  &0.000  &0.294  &0.412  &0.343 \\
    \bottomrule
    \end{tabular}
  \label{tab7}
\end{table}

\begin{table}[htbp]
  \centering
  \normalsize
  \setlength\tabcolsep{3pt}
  \caption{Performances with different frames as point-level labels on the CAS(ME)$^2$ dataset. }
    \begin{tabular}{l|cccccc}
    \toprule
    Supervision   &F1(0.5)&F1(1.0)&F1(p)&REC&PRE&F1 \\
    \midrule
    Apex     &\textbf{0.235} &\textbf{0.112}  &\textbf{0.100}  &0.302  &\textbf{0.489}  &\textbf{0.374} \\
    Random   &0.203 &0.102 &0.023 &\textbf{0.305} &0.467 &0.366 \\
    \bottomrule
    \end{tabular}
  \label{tab8}
\end{table}

\begin{table*}[htbp]
  \centering
  \small 
  \setlength\tabcolsep{5pt}
  \caption{Comparison with the state-of-the-art models on the CAS(ME)$^2$ dataset. We also cover the methods under video-level and frame-level supervision 
  for reference. The best performances for each of the three groups are highlighted in bold. }
    \begin{tabular}{c|c|cccccc}
    \toprule
    Supervision & Method &F1(0.5)&F1(1.0)&F1(p)&REC&PRE&F1\\
    \midrule
    \multirow{8}[2]{*}{\makecell{Frame-level \\ (Full)}}  &He et al. (2020)\cite{he2020spotting}     &   -  &   -  &0.008 &0.020 &0.364 &0.038 \\
                                                          &Zhang et al. (2020)\cite{zhang2020spatio} &   -  &   -  &0.055 &0.085 &0.406 &0.140 \\
                                                          &MESNet (2021) \cite{wang2021mesnet}       &   -  &   -  &  -   &  -   &   -  &0.036 \\
                                                          &Yap et al. (2021) \cite{yap20213dcnn}     &   -  &   -  &0.012 &  -   &   -  &0.030 \\
                                                          &LSSNet (2021) \cite{yu2021lssnet}         &   -  &   -  &0.063 &  -   &   -  &0.327 \\
                                                          &He et al. (2021) \cite{yuhong2021research}&   -  &   -  &\textbf{0.197} &  -   &   -  &0.343 \\
                                                          &MTSN (2022) \cite{liong2022mtsn}          &   -  &   -  &0.081 &0.342 &0.385 &0.362 \\
                                                          &Zhao et al. (2022) \cite{zhao2022rethink} &   -  &   -  &   -  &   -  &   -  &0.403 \\
                                                          &LGSNet (2023)\cite{yu2023lgsnet}          &   -  &   -  &   -  &\textbf{0.367} &\textbf{0.630} &\textbf{0.464} \\
    \midrule
    \multirow{4}[2]{*}{\makecell{Video-level \\ (Weak)}}  &HAM-Net (2021) \cite{islam2021hybrid}     &0.010 &0.007 &0.000 &0.042 &0.090 &0.057 \\
                                                          &CO$_2$-Net (2021) \cite{hong2021cross}    &0.057 &0.031 &0.000 &0.095 &0.153 &0.117 \\
                                                          &FTCL (2022) \cite{gao2022fine}            &0.092 &0.022 &0.000 &0.048 &0.070 &0.057 \\
                                                          &MC-WES \cite{yu2023weaklysupervised}      &\textbf{0.167} &\textbf{0.108} &\textbf{0.169} &\textbf{0.266} &\textbf{0.415} &\textbf{0.324} \\
    
    \midrule
    \multirow{3}[2]{*}{\makecell{Point-level \\ (Weak)}}  &SF-Net (2020) \cite{ma2020sf}             &0.139 &0.089 &0.096 &0.270 &0.334 &0.298 \\
                                                          &LACP (2021)  \cite{lee2021learning}       &0.019 &0.016 &0.020 &0.120 &0.091 &0.103 \\
                                                          &\textbf{PWES}                             &\textbf{0.203} &\textbf{0.102} &\textbf{0.023} &\textbf{0.305} &\textbf{0.467} &\textbf{0.366} \\
    \bottomrule
    \end{tabular}
  \label{tab9}
\end{table*}

\begin{table*}[htbp]
  \centering
  \small 
  \setlength\tabcolsep{5pt}
  \caption{Comparison with the fully-supervised state-of-the-art models on the SAMM-LV dataset.}
    \begin{tabular}{c|c|cccccc}
    \toprule
    Supervision & Method &F1(0.5)&F1(1.0)&F1(p)&REC&PRE&F1 \\
    \midrule
    \multirow{8}[2]{*}{\makecell{Frame-level \\ (Full)}}  &He et al.  (2020) \cite{he2020spotting}   &   -  &   -  &0.036 &0.029 &0.101 &0.045 \\
                                                          &Zhang et al. (2020) \cite{zhang2020spatio}&   -  &   -  &0.073 &0.079 &0.136 &0.100 \\
                                                          &MESNet (2021) \cite{wang2021mesnet}       &   -  &   -  &   -  &   -  &  -   &0.088 \\
                                                          &Yap et al. (2021) \cite{yap20213dcnn}     &   -  &   -  &0.044 &   -  &  -   &0.119 \\
                                                          &LSSNet (2021) \cite{yu2021lssnet}         &   -  &   -  &\textbf{0.218} &   -  &  -   &0.290 \\
                                                          &He et al. (2021) \cite{yuhong2021research}&   -  &   -  &0.216 &   -  &  -   &0.364 \\
                                                          &MTSN (2022) \cite{liong2022mtsn}          &   -  &   -  &0.088 &0.260 &0.319 &0.287 \\
                                                          &Zhao et al. (2022) \cite{zhao2022rethink} &   -  &   -  &   -  &   -  &   -  &0.386 \\
                                                          &LGSNet (2023)\cite{yu2023lgsnet}          &   -  &   -  &   -  &\textbf{0.355} &\textbf{0.429} &\textbf{0.388} \\
    \midrule
    \multirow{4}[2]{*}{\makecell{Video-level \\ (Weak)}}  &HAM-Net (2021) \cite{islam2021hybrid}     &0.113 &\textbf{0.060} &0.028 &0.150 &0.113 &0.129\\
                                                          &CO$_2$-Net (2021) \cite{hong2021cross}    &0.111 &0.058 &0.039 &0.230 &0.148 &0.181 \\
                                                          &FTCL (2022) \cite{gao2022fine}            &0.116 &0.048 &0.004 &0.142 &0.138 &0.140 \\
                                                          &MC-WES (2023) \cite{yu2023weaklysupervised}      &\textbf{0.135} &0.055 &\textbf{0.135} &\textbf{0.263} &\textbf{0.178} &\textbf{0.212} \\
    \midrule
    \multirow{3}[2]{*}{\makecell{Point-level \\ (Weak)}}  &SF-Net (2020) \cite{ma2020sf}             &0.088 &0.075 &0.036 &0.132 &\textbf{0.253} &0.174 \\
                                                          &LACP (2021)  \cite{lee2021learning}       &0.097 &0.049 &\textbf{0.096} &0.164 &0.151 &0.157 \\ 
                                                          &\textbf{PWES}                             &\textbf{0.310} &\textbf{0.077} &0.000 &\textbf{0.286} &0.226 &\textbf{0.252} \\
    \bottomrule
    \end{tabular}
  \label{tab10}
\end{table*}

\begin{table*}[htbp]
  \centering
  \small 
  \setlength\tabcolsep{5pt}
  \caption{Performances on the CAS(ME)$^3$ dataset.}
    \begin{tabular}{c|l|cccccc}
    \toprule
    Supervision & Method &F1-ME(0.5)&F1-ME(1.0)&F1-ME(p)&Recall&Precision&F1-score \\
    \midrule
    \multirow{3}[2]{*}{Full}                              &SP-FD (2020)\cite{zhang2020spatio}    &0.010 &0.010 &  -   &  -   &  -   &  -  \\
                                                          &LSSNet (2021)\cite{yu2021lssnet}      &0.065 &0.065 &  -   &  -   &   -  &  -  \\
                                                          &LGSNet (2023)\cite{yu2023lgsnet}      &\textbf{0.171} &\textbf{0.136} &\textbf{0.099} &\textbf{0.292} &\textbf{0.196} &\textbf{0.235} \\

    \midrule
    \multirow{4}[2]{*}{\makecell{Video-level \\ (Weak)}}  &HAM-Net (2021)\cite{islam2021hybrid}  &0.008 &0.006 &0.000 &0.098 &0.030 &0.046 \\
                                                          &CO$_2$-Net (2021)\cite{hong2021cross} &0.037 &0.018 &0.000 &0.118 &0.050 &0.070 \\
                                                          &FTCL (2022) \cite{gao2022fine}        &0.014 &0.012 &0.000 &0.106 &0.034 &0.052 \\
                                                          &MC-WES (2023) \cite{yu2023weaklysupervised}  &\textbf{0.048} &\textbf{0.022} &0.000 &\textbf{0.141} &\textbf{0.060} &\textbf{0.084} \\
    \midrule
    \multirow{3}[2]{*}{\makecell{Point-level \\ (Weak)}}  &SF-Net (2020) \cite{ma2020sf}         &0.010 &0.006 &\textbf{0.017} &0.075 &0.036 &0.049 \\
                                                          &LACP (2021)  \cite{lee2021learning}   &0.051 &0.023 &0.000 &0.111 &0.078 &0.091 \\
                                                          &\textbf{PWES}                         &\textbf{0.093} &\textbf{0.047} &0.000 &\textbf{0.171} &\textbf{0.158} &\textbf{0.164} \\
    \bottomrule
    \end{tabular}
  \label{tab11}
\end{table*}

\noindent \textbf{Number of background region-level vector.}
The distribution of background snippets is denser compared with that of foreground snippets \cite{xu2016heterogeneous, qu2017cas, yap2020samm}, requiring us to employ 
a higher number of background region-level vectors for representing detailed background features. Here we investigate the impact of the number of background 
region-level vectors, as outlined in Table \ref{tab7}. The results indicate that the too low numbers inhibit the recall and precision of overall expression 
spotting, while the excessively large numbers noticeably reduce precision. Conversely, for ME spotting, increasing the number of vectors improves the results. Essentially, 
we posit that a greater number of background region-level vectors facilitate the learning of finer-grained features, enabling the better identification of short intervals 
associated with MEs. On the other hand, insufficient background region-level vectors hinder model performance by impeding the acquisition of effective representations.

\subsection{Different Point-level Supervision}
Existing key-frame methods \cite{zhang2018smeconvnet, liong2021shallow} commonly rely on apex frames for frame-level spotting of MEs and MaEs. However, the annotation of 
apex frames is inherently tied to the annotation of onset and offset frames, making it a fully-supervised learning setting. In contrast, our arbitrary single-frame supervision 
only requires coarse labels in model training, bringing significant cost reduction. To validate the effectiveness of our arbitrary single-frame supervision, we also 
train the model using apex-frame supervision under the same settings. \textcolor{black}{For simplicity, we refer to the arbitrary single-frame supervision method as the random point-level method
and the apex-frame supervision method as the apex point-level method, as shown in Table \ref{tab8}. Comparing the F1-scores from Table \ref{tab8}, it is evident that the 
F1-score of the random point-level method closely aligns with that of the apex point-level method.}
Moreover, we observe that the recall rates of the two supervision methods are similar, while the precision of apex point-level method shows a notable decrease of 
2.2\% compared with our random point-level method. Regarding ME spotting, we find that F1(0.5) and F1(1.0) of the two supervision methods are basically similar, 
except for F1(P) where the random point-level method yields a significantly lower result than that of the apex point-level method by 8.9\%. 
The results demonstrate that our weak supervision approach achieves the performance close to that of the apex-frame based full supervision, in terms of MEs and the overall proposal sets.
 
\subsection{Comparison with State-of-the-art Methods}
To the best of our knowledge, our work is the first to employ point-level annotation to achieve weakly-supervised frame-level expression spotting. Hence, we can only compare our approach with 
those fully-supervised and video-level weakly-supervised methods. In addition, we also conduct comparison with recent point-level weakly-supervised methods originally designed for WTAL tasks on 
the same datasets including SF-Net \cite{ma2020sf} and LACP \cite{lee2021learning}. 

Table \ref{tab9} and Table \ref{tab10} present detailed comparisons on the CAS(ME)$^2$ and SAMM-LV datasets, respectively. The results demonstrate that incorporating arbitrary single-frame 
(point-level) supervision information into the video-level methods using our proposed MPLG and DFCL algorithms leads to a significant improvement in performance 
on the CAS(ME)$^2$ dataset compared with MC-WES \cite{yu2023weaklysupervised}. 
Specifically, our PWES shows a substantial increase of 3.9\% in overall recall, 5.2\% in overall precision, and 4.4\% in overall F1-score. Similarly, on the SAMM-LV dataset, our PWES obtains 
notable improvements of 2.3\% in overall recall, 4.8\% in overall precision, and 4.0\% in overall F1-score. In addition, we also find that our PWES distinctly outperforms other existing point-level 
weakly-supervised methods on the two datasets.

However, as for the ME spotting, we notice that these findings solidify the effectiveness of our point-level method and highlight the almost equivalent performance of our weakly-supervised approach 
to the fully-supervised approaches. On the CAS(ME)$^2$ dataset, there is a 14.6\% decline in F1(P), a 3.6\% increase in F1(0.5), and a 0.6\% decline in F1(1.0). 
On the SAMM-LV dataset, there is a 13.5\% decline in F1(P), a 17.5\% increase in F1(0.5), and a 2.2\% increase in F1(1.0). One immediate reason for this disparity is that MEs have shorter durations, 
typically spanning only three snippets in a video (with a frame rate of 30 FPS and a snippet length of 8 frames). Consequently, learning ME features becomes quite challenging when the timestamps 
of random frames deviate significantly from the apex frames. These findings are also supported by the results presented in Table \ref{tab8}.

\textcolor{black}{Given the limited sample sizes in CAS(ME)$^2$ and SAMM-LV in comparison to other datasets in computer vision, we extend our evaluation to a more extensive dataset, CAS(ME)$^3$, comprising 956 videos.
Table \ref{tab11} demonstrates PWES's substantial improvements compared to recent point-level weakly-supervised methods designed for WTAL tasks with at least a 7.3\% increase in F1-score, a 6.0\% rise in recall, and 
an 8.0\% improvement in precision. Notably, in ME spotting, F1-ME(P) remains consistently with other methods, while both F1-ME(0.5) and F1-ME(1.0) show significant enhancements, validating the 
efficacy of our approach on larger datasets.}

\section{Conclusion}
\label{sec5}
In this paper, we introduce a novel weakly-supervised approach that utilizes randomly selected single frames from each ground truth intervals as training labels. This strategy not only 
effectively reduces the labor and time required for annotation, but also significantly improves the performance of video-level methods.
Additionally, we propose a new algorithm called MPLG to generate more reliable pseudo labels by merging class-specific probabilities, attention scores, 
current video features, and point-level labels. This algorithm tackles the limitations of previous pseudo label generation methods, which often result in models localizing 
high-confidence contextual backgrounds and discarding low-confidence foregrounds, leading to incorrect expression boundaries and fragmentary predictions. In addition, 
we introduce the DFCL algorithm to capture global features across the entire dataset, enhancing feature similarity for the same categories and feature variability 
for different categories. Extensive experiments conducted on the CAS(ME)$^2$, CAS(ME)$^3$ and SAMM-LV datasets validate the effectiveness of our proposed method.

It should be noted that our method struggles to spot MEs in the optimal overall proposal set, which is the main reason for the slightly lower F1(p) scores. Due to the inherent properties 
of MEs, their distribution is sparser than that of MaEs. This imbalanced distribution significantly impacts the learning of 
MEs and leads to relatively poor testing results. Therefore, in future work, we plan to explore strategies to effectively deal with the category imbalance.


%

\ifCLASSOPTIONcompsoc
  \section*{Acknowledgments}
\else
  \section*{Acknowledgment}
\fi

This work was supported by STI2030-Major Projects (\#2022ZD0204600).

\ifCLASSOPTIONcaptionsoff
  \newpage
\fi



%



\bibliographystyle{IEEEtran}
\bibliography{ref}

\begin{thebibliography}{10}
\providecommand{\url}[1]{#1}
\csname url@samestyle\endcsname
\providecommand{\newblock}{\relax}
\providecommand{\bibinfo}[2]{#2}
\providecommand{\BIBentrySTDinterwordspacing}{\spaceskip=0pt\relax}
\providecommand{\BIBentryALTinterwordstretchfactor}{4}
\providecommand{\BIBentryALTinterwordspacing}{\spaceskip=\fontdimen2\font plus
\BIBentryALTinterwordstretchfactor\fontdimen3\font minus
  \fontdimen4\font\relax}
\providecommand{\BIBforeignlanguage}[2]{{%
\expandafter\ifx\csname l@#1\endcsname\relax
\typeout{** WARNING: IEEEtran.bst: No hyphenation pattern has been}%
\typeout{** loaded for the language `#1'. Using the pattern for}%
\typeout{** the default language instead.}%
\else
\language=\csname l@#1\endcsname
\fi
#2}}
\providecommand{\BIBdecl}{\relax}
\BIBdecl

\bibitem{ekman1969nonverbal}
P.~Ekman and W.~V. Friesen, ``Nonverbal leakage and clues to deception,''
  \emph{Psychiatry}, vol.~32, no.~1, pp. 88--106, 1969.

\bibitem{ekman2003darwin}
P.~Ekman, ``Darwin, deception, and facial expression,'' \emph{Ann. New York
  Acad. Sci.}, vol. 1000, no.~1, pp. 205--221, 2003.

\bibitem{ekman2009lie}
P.~\vspace{0mm} Ekman, ``Lie catching and microexpressions,'' \emph{The
  philosophy of deception}, vol.~1, no.~2, p.~5, 2009.

\bibitem{yu2023weaklysupervised}
W.-W. Yu, K.-F. Yang, H.-M. Yan, and Y.-J. Li, ``Weakly-supervised micro- and
  macro-expression spotting based on multi-level consistency,'' \emph{arXiv
  preprint arXiv:2305.02734}, 2023.

\bibitem{xie2019adaptive}
W.~Xie, L.~Shen, and J.~Duan, ``Adaptive weighting of handcrafted feature
  losses for facial expression recognition,'' \emph{{IEEE} Trans. Cybern.},
  vol.~51, no.~5, pp. 2787--2800, 2021.

\bibitem{du2014compound}
S.~Du, Y.~Tao, and A.~M. Mart{\'{\i}}nez, ``Compound facial expressions of
  emotion,'' \emph{Proc. Natl. Acad. Sci. {USA}}, vol. 111, no.~15, pp.
  E1454--E1462, 2014.

\bibitem{li2017reliable}
S.~Li, W.~Deng, and J.~Du, ``Reliable crowdsourcing and deep
  locality-preserving learning for expression recognition in the wild,'' in
  \emph{{CVPR}}.\hskip 1em plus 0.5em minus 0.4em\relax {IEEE} Computer
  Society, 2017, pp. 2584--2593.

\bibitem{liang2020fine}
L.~Liang, C.~Lang, Y.~Li, S.~Feng, and J.~Zhao, ``Fine-grained facial
  expression recognition in the wild,'' \emph{{IEEE} Trans. Inf. Forensics
  Secur.}, vol.~16, pp. 482--494, 2021.

\bibitem{davison2016samm}
A.~K. Davison, C.~Lansley, N.~Costen, K.~Tan, and M.~H. Yap, ``{SAMM:} {A}
  spontaneous micro-facial movement dataset,'' \emph{{IEEE} Trans. Affect.
  Comput.}, vol.~9, no.~1, pp. 116--129, 2018.

\bibitem{li2013spontaneous}
X.~Li, T.~Pfister, X.~Huang, G.~Zhao, and M.~Pietik{\"{a}}inen, ``A spontaneous
  micro-expression database: Inducement, collection and baseline,'' in
  \emph{{FG}}.\hskip 1em plus 0.5em minus 0.4em\relax {IEEE} Computer Society,
  2013, pp. 1--6.

\bibitem{yan2013casme}
W.~Yan, Q.~Wu, Y.~Liu, S.~Wang, and X.~Fu, ``{CASME} database: {A} dataset of
  spontaneous micro-expressions collected from neutralized faces,'' in
  \emph{{FG}}.\hskip 1em plus 0.5em minus 0.4em\relax {IEEE} Computer Society,
  2013, pp. 1--7.

\bibitem{yan2014casme}
W.~Yan, X.~Li, S.~Wang, G.~Zhao, Y.~Liu, Y.~Chen, and X.~Fu, ``{CASME II}: {A}n
  improved spontaneous micro-expression database and the baseline evaluation,''
  \emph{PloS one}, vol.~9, no.~1, p. e86041, 2014.

\bibitem{zhao2011facial}
G.~Zhao, X.~Huang, M.~Taini, S.~Z. Li, and M.~Pietik{\"{a}}inen, ``Facial
  expression recognition from near-infrared videos,'' \emph{Image Vis.
  Comput.}, vol.~29, no.~9, pp. 607--619, 2011.

\bibitem{kossaifi2019sewa}
J.~Kossaifi, R.~Walecki, Y.~Panagakis, J.~Shen, M.~Schmitt, F.~Ringeval,
  J.~Han, V.~Pandit, A.~Toisoul, B.~W. Schuller, K.~Star, E.~Hajiyev, and
  M.~Pantic, ``{SEWA} {DB:} {A} rich database for audio-visual emotion and
  sentiment research in the wild,'' \emph{{IEEE} Trans. Pattern Anal. Mach.
  Intell.}, vol.~43, no.~3, pp. 1022--1040, 2021.

\bibitem{kollias2020analysing}
D.~Kollias, A.~Schulc, E.~Hajiyev, and S.~Zafeiriou, ``Analysing affective
  behavior in the first {ABAW} 2020 competition,'' in \emph{{FG}}.\hskip 1em
  plus 0.5em minus 0.4em\relax {IEEE}, 2020, pp. 637--643.

\bibitem{qu2017cas}
F.~Qu, S.~Wang, W.~Yan, H.~Li, S.~Wu, and X.~Fu, ``{CAS(ME)\({}^{\mbox{2}}\)}:
  {A} database for spontaneous macro-expression and micro-expression spotting
  and recognition,'' \emph{{IEEE} Trans. Affect. Comput.}, vol.~9, no.~4, pp.
  424--436, 2018.

\bibitem{li2022cas}
J.~Li, Z.~Dong, S.~Lu, S.-J. Wang, W.-J. Yan, Y.~Ma, Y.~Liu, C.~Huang, and
  X.~Fu, ``{CAS(ME)}\textsuperscript{3}: A third generation facial spontaneous
  micro-expression database with depth information and high ecological
  validity,'' \emph{{IEEE} Trans. Pattern Anal. Mach. Intell.}, 2022.

\bibitem{yap2020samm}
C.~H. Yap, C.~Kendrick, and M.~H. Yap, ``{SAMM} long videos: {A} spontaneous
  facial micro- and macro-expressions dataset,'' in \emph{{FG}}.\hskip 1em plus
  0.5em minus 0.4em\relax {IEEE}, 2020, pp. 771--776.

\bibitem{ben2021video}
X.~Ben, Y.~Ren, J.~Zhang, S.~Wang, K.~Kpalma, W.~Meng, and Y.~Liu,
  ``Video-based facial micro-expression analysis: {A} survey of datasets,
  features and algorithms,'' \emph{{IEEE} Trans. Pattern Anal. Mach. Intell.},
  vol.~44, no.~9, pp. 5826--5846, 2022.

\bibitem{wang2021mesnet}
S.~Wang, Y.~He, J.~Li, and X.~Fu, ``{MESNet}: {A} convolutional neural network
  for spotting multi-scale micro-expression intervals in long videos,''
  \emph{{IEEE} Trans. Image Process.}, vol.~30, pp. 3956--3969, 2021.

\bibitem{lu2022more}
S.~Lu, J.~Li, Y.~Wang, Z.~Dong, S.-J. Wang, and X.~Fu, ``{A} more objective
  quantification of micro-expression intensity through facial
  electromyography,'' in \emph{Proceedings of the 2nd Workshop on Facial
  Micro-Expression: Advanced Techniques for Multi-Modal Facial Expression
  Analysis}, 2022, pp. 11--17.

\bibitem{ekman1993facial}
P.~Ekman, ``Facial expression and emotion,'' \emph{American Psychologist},
  vol.~48, pp. 384--392, 1993.

\bibitem{esposito2007amount}
A.~Esposito, ``The amount of information on emotional states conveyed by the
  verbal and nonverbal channels: Some perceptual data,'' in \emph{Progress in
  nonlinear speech processing}.\hskip 1em plus 0.5em minus 0.4em\relax
  Springer, 2007, pp. 249--268.

\bibitem{he2022micro}
Y.~He, ``Research on micro-expression spotting method based on optical flow
  features,'' in \emph{{ACM} Multimedia}.\hskip 1em plus 0.5em minus
  0.4em\relax {ACM}, 2021, pp. 4803--4807.

\bibitem{yu2021lssnet}
W.~Yu, J.~Jiang, and Y.~Li, ``Lssnet: {A} two-stream convolutional neural
  network for spotting macro- and micro-expression in long videos,'' in
  \emph{{ACM} Multimedia}.\hskip 1em plus 0.5em minus 0.4em\relax {ACM}, 2021,
  pp. 4745--4749.

\bibitem{guo2023micro}
X.~Guo, X.~Zhang, L.~Li, and Z.~Xia, ``Micro-expression spotting with
  multi-scale local transformer in long videos,'' \emph{Pattern Recognition
  Letters}, 2023.

\bibitem{yu2023lgsnet}
W.-W. Yu, J.~Jiang, K.-F. Yang, H.-M. Yan, and Y.-J. Li, ``{LGSN}et: {A}
  two-stream network for micro-and macro-expression spotting with background
  modeling,'' \emph{IEEE Transactions on Affective Computing}, 2023.

\bibitem{islam2021hybrid}
A.~Islam, C.~Long, and R.~J. Radke, ``A hybrid attention mechanism for
  weakly-supervised temporal action localization,'' in \emph{{AAAI}}.\hskip 1em
  plus 0.5em minus 0.4em\relax {AAAI} Press, 2021, pp. 1637--1645.

\bibitem{hong2021cross}
F.~Hong, J.~Feng, D.~Xu, Y.~Shan, and W.~Zheng, ``Cross-modal consensus network
  for weakly supervised temporal action localization,'' in \emph{{ACM}
  Multimedia}.\hskip 1em plus 0.5em minus 0.4em\relax {ACM}, 2021, pp.
  1591--1599.

\bibitem{paul2018w}
S.~Paul, S.~Roy, and A.~K. Roy{-}Chowdhury, ``{W-TALC:} weakly-supervised
  temporal activity localization and classification,'' in \emph{{ECCV} {(4)}},
  ser. Lecture Notes in Computer Science, vol. 11208.\hskip 1em plus 0.5em
  minus 0.4em\relax Springer, 2018, pp. 588--607.

\bibitem{wang2017untrimmednets}
L.~Wang, Y.~Xiong, D.~Lin, and L.~V. Gool, ``Untrimmednets for weakly
  supervised action recognition and detection,'' in \emph{{CVPR}}.\hskip 1em
  plus 0.5em minus 0.4em\relax {IEEE} Computer Society, 2017, pp. 6402--6411.

\bibitem{nguyen2018weakly}
P.~Nguyen, T.~Liu, G.~Prasad, and B.~Han, ``Weakly supervised action
  localization by sparse temporal pooling network,'' in \emph{{CVPR}}.\hskip
  1em plus 0.5em minus 0.4em\relax Computer Vision Foundation / {IEEE} Computer
  Society, 2018, pp. 6752--6761.

\bibitem{nguyen2019weakly}
P.~X. Nguyen, D.~Ramanan, and C.~C. Fowlkes, ``Weakly-supervised action
  localization with background modeling,'' in \emph{{ICCV}}.\hskip 1em plus
  0.5em minus 0.4em\relax {IEEE}, 2019, pp. 5501--5510.

\bibitem{su2018cascaded}
H.~Su, X.~Zhao, and T.~Lin, ``Cascaded pyramid mining network for weakly
  supervised temporal action localization,'' in \emph{{ACCV} {(2)}}, ser.
  Lecture Notes in Computer Science, vol. 11362.\hskip 1em plus 0.5em minus
  0.4em\relax Springer, 2018, pp. 558--574.

\bibitem{ma2020sf}
F.~Ma, L.~Zhu, Y.~Yang, S.~Zha, G.~Kundu, M.~Feiszli, and Z.~Shou, ``Sf-net:
  Single-frame supervision for temporal action localization,'' in \emph{{ECCV}
  {(4)}}, ser. Lecture Notes in Computer Science, vol. 12349.\hskip 1em plus
  0.5em minus 0.4em\relax Springer, 2020, pp. 420--437.

\bibitem{lee2021learning}
P.~Lee and H.~Byun, ``Learning action completeness from points for
  weakly-supervised temporal action localization,'' in \emph{{ICCV}}.\hskip 1em
  plus 0.5em minus 0.4em\relax {IEEE}, 2021, pp. 13\,628--13\,637.

\bibitem{fu2022compact}
J.~Fu, J.~Gao, and C.~Xu, ``Compact representation and reliable classification
  learning for point-level weakly-supervised action localization,''
  \emph{{IEEE} Trans. Image Process.}, vol.~31, pp. 7363--7377, 2022.

\bibitem{huang2022weakly}
L.~Huang, L.~Wang, and H.~Li, ``Weakly supervised temporal action localization
  via representative snippet knowledge propagation,'' in \emph{{CVPR}}.\hskip
  1em plus 0.5em minus 0.4em\relax {IEEE}, 2022, pp. 3262--3271.

\bibitem{carreira2017quo}
J.~Carreira and A.~Zisserman, ``Quo vadis, action recognition? {A} new model
  and the kinetics dataset,'' in \emph{{CVPR}}.\hskip 1em plus 0.5em minus
  0.4em\relax {IEEE} Computer Society, 2017, pp. 4724--4733.

\bibitem{Dmaron1997framework}
O.~Maron and T.~Lozano{-}P{\'{e}}rez, ``A framework for multiple-instance
  learning,'' in \emph{{NIPS}}.\hskip 1em plus 0.5em minus 0.4em\relax The
  {MIT} Press, 1997, pp. 570--576.

\bibitem{pan2020local}
H.~Pan, L.~Xie, and Z.~Wang, ``Local bilinear convolutional neural network for
  spotting macro- and micro-expression intervals in long video sequences,'' in
  \emph{{FG}}.\hskip 1em plus 0.5em minus 0.4em\relax {IEEE}, 2020, pp.
  749--753.

\bibitem{zhang2018smeconvnet}
Z.~Zhang, T.~Chen, H.~Meng, G.~Liu, and X.~Fu, ``{SMEConvNet}: {A}
  convolutional neural network for spotting spontaneous facial micro-expression
  from long videos,'' \emph{{IEEE} Access}, vol.~6, pp. 71\,143--71\,151, 2018.

\bibitem{yap20213dcnn}
C.~H. Yap, M.~H. Yap, A.~K. Davison, C.~Kendrick, J.~Li, S.~Wang, and
  R.~Cunningham, ``3d-cnn for facial micro- and macro-expression spotting on
  long video sequences using temporal oriented reference frame,'' pp.
  7016--7020, 2022.

\bibitem{liong2021shallow}
G.~Liong, J.~See, and L.~Wong, ``Shallow optical flow three-stream {CNN} for
  macro- and micro-expression spotting from long videos,'' in
  \emph{{ICIP}}.\hskip 1em plus 0.5em minus 0.4em\relax {IEEE}, 2021, pp.
  2643--2647.

\bibitem{sun2019two}
B.~Sun, S.~Cao, J.~He, and L.~Yu, ``Two-stream attention-aware network for
  spontaneous micro-expression movement spotting,'' in \emph{{ICSESS}}.\hskip
  1em plus 0.5em minus 0.4em\relax {IEEE}, 2019, pp. 702--705.

\bibitem{tran2019dense}
T.~Tran, Q.~Vo, X.~Hong, and G.~Zhao, ``Dense prediction for micro-expression
  spotting based on deep sequence model,'' \emph{Electronic Imaging}, vol.
  2019, no.~8, pp. 401--1, 2019.

\bibitem{verburg2019micro}
M.~Verburg and V.~Menkovski, ``Micro-expression detection in long videos using
  optical flow and recurrent neural networks,'' in \emph{{FG}}.\hskip 1em plus
  0.5em minus 0.4em\relax {IEEE}, 2019, pp. 1--6.

\bibitem{lin2017single}
T.~Lin, X.~Zhao, and Z.~Shou, ``Single shot temporal action detection,'' in
  \emph{{ACM} Multimedia}.\hskip 1em plus 0.5em minus 0.4em\relax {ACM}, 2017,
  pp. 988--996.

\bibitem{yang2020revisiting}
L.~Yang, H.~Peng, D.~Zhang, J.~Fu, and J.~Han, ``Revisiting anchor mechanisms
  for temporal action localization,'' \emph{{IEEE} Trans. Image Process.},
  vol.~29, pp. 8535--8548, 2020.

\bibitem{zhao2017temporal}
Y.~Zhao, Y.~Xiong, L.~Wang, Z.~Wu, X.~Tang, and D.~Lin, ``Temporal action
  detection with structured segment networks,'' in \emph{{ICCV}}.\hskip 1em
  plus 0.5em minus 0.4em\relax {IEEE} Computer Society, 2017, pp. 2933--2942.

\bibitem{zhang2020asfd}
J.~Li, B.~Zhang, Y.~Wang, Y.~Tai, Z.~Zhang, C.~Wang, J.~Li, X.~Huang, and
  Y.~Xia, ``{ASFD:} automatic and scalable face detector,'' in \emph{{ACM}
  Multimedia}.\hskip 1em plus 0.5em minus 0.4em\relax {ACM}, 2021, pp.
  2139--2147.

\bibitem{ju2020point}
C.~Ju, P.~Zhao, Y.~Zhang, Y.~Wang, and Q.~Tian, ``Point-level temporal action
  localization: Bridging fully-supervised proposals to weakly-supervised
  losses,'' \emph{CoRR}, vol. abs/2012.08236, 2020.

\bibitem{ju2021divide}
C.~Ju, P.~Zhao, S.~Chen, Y.~Zhang, Y.~Wang, and Q.~Tian, ``Divide and conquer
  for single-frame temporal action localization,'' in \emph{{ICCV}}.\hskip 1em
  plus 0.5em minus 0.4em\relax {IEEE}, 2021, pp. 13\,435--13\,444.

\bibitem{narayan20193c}
S.~Narayan, H.~Cholakkal, F.~S. Khan, and L.~Shao, ``3c-net: Category count and
  center loss for weakly-supervised action localization,'' in
  \emph{{ICCV}}.\hskip 1em plus 0.5em minus 0.4em\relax {IEEE}, 2019, pp.
  8678--8686.

\bibitem{shi2020weakly}
B.~Shi, Q.~Dai, Y.~Mu, and J.~Wang, ``Weakly-supervised action localization by
  generative attention modeling,'' in \emph{{CVPR}}.\hskip 1em plus 0.5em minus
  0.4em\relax Computer Vision Foundation / {IEEE}, 2020, pp. 1006--1016.

\bibitem{liu2019completeness}
D.~Liu, T.~Jiang, and Y.~Wang, ``Completeness modeling and context separation
  for weakly supervised temporal action localization,'' in \emph{{CVPR}}.\hskip
  1em plus 0.5em minus 0.4em\relax Computer Vision Foundation / {IEEE}, 2019,
  pp. 1298--1307.

\bibitem{lee2020background}
P.~Lee, Y.~Uh, and H.~Byun, ``Background suppression network for
  weakly-supervised temporal action localization,'' in \emph{{AAAI}}.\hskip 1em
  plus 0.5em minus 0.4em\relax {AAAI} Press, 2020, pp. 11\,320--11\,327.

\bibitem{pardo2021refineloc}
A.~Pardo, H.~Alwassel, F.~C. Heilbron, A.~K. Thabet, and B.~Ghanem,
  ``{RefineLoc}: Iterative refinement for weakly-supervised action
  localization,'' in \emph{{WACV}}.\hskip 1em plus 0.5em minus 0.4em\relax
  {IEEE}, 2021, pp. 3318--3327.

\bibitem{luo2020weakly}
Z.~Luo, D.~Guillory, B.~Shi, W.~Ke, F.~Wan, T.~Darrell, and H.~Xu,
  ``Weakly-supervised action localization with expectation-maximization
  multi-instance learning,'' in \emph{{ECCV} {(29)}}, vol. 12374.\hskip 1em
  plus 0.5em minus 0.4em\relax Springer, 2020, pp. 729--745.

\bibitem{he2022asm}
B.~He, X.~Yang, L.~Kang, Z.~Cheng, X.~Zhou, and A.~Shrivastava, ``Asm-loc:
  Action-aware segment modeling for weakly-supervised temporal action
  localization,'' in \emph{{CVPR}}.\hskip 1em plus 0.5em minus 0.4em\relax
  {IEEE}, 2022, pp. 13\,915--13\,925.

\bibitem{moltisanti2019action}
D.~Moltisanti, S.~Fidler, and D.~Damen, ``Action recognition from single
  timestamp supervision in untrimmed videos,'' in \emph{{CVPR}}.\hskip 1em plus
  0.5em minus 0.4em\relax Computer Vision Foundation / {IEEE}, 2019, pp.
  9915--9924.

\bibitem{yang2021uncertainty}
W.~Yang, T.~Zhang, X.~Yu, Q.~Tian, Y.~Zhang, and F.~Wu, ``Uncertainty guided
  collaborative training for weakly supervised temporal action detection,'' in
  \emph{{CVPR}}.\hskip 1em plus 0.5em minus 0.4em\relax Computer Vision
  Foundation / {IEEE}, 2021, pp. 53--63.

\bibitem{khosla2020supervised}
P.~Khosla, P.~Teterwak, C.~Wang, A.~Sarna, Y.~Tian, P.~Isola, A.~Maschinot,
  C.~Liu, and D.~Krishnan, ``Supervised contrastive learning,'' in
  \emph{NeurIPS}, 2020.

\bibitem{zheng2021weakly}
M.~Zheng, F.~Wang, S.~You, C.~Qian, C.~Zhang, X.~Wang, and C.~Xu, ``Weakly
  supervised contrastive learning,'' in \emph{{ICCV}}.\hskip 1em plus 0.5em
  minus 0.4em\relax {IEEE}, 2021, pp. 10\,022--10\,031.

\bibitem{chen2020simple}
T.~Chen, S.~Kornblith, M.~Norouzi, and G.~E. Hinton, ``A simple framework for
  contrastive learning of visual representations,'' in \emph{{ICML}}, ser.
  Proceedings of Machine Learning Research, vol. 119.\hskip 1em plus 0.5em
  minus 0.4em\relax {PMLR}, 2020, pp. 1597--1607.

\bibitem{he2020momentum}
K.~He, H.~Fan, Y.~Wu, S.~Xie, and R.~B. Girshick, ``Momentum contrast for
  unsupervised visual representation learning,'' in \emph{{CVPR}}.\hskip 1em
  plus 0.5em minus 0.4em\relax Computer Vision Foundation / {IEEE}, 2020, pp.
  9726--9735.

\bibitem{kaya2019deep}
M.~Kaya and H.~S. Bilge, ``Deep metric learning: {A} survey,'' \emph{Symmetry},
  vol.~11, no.~9, p. 1066, 2019.

\bibitem{wang2021exploring}
W.~Wang, T.~Zhou, F.~Yu, J.~Dai, E.~Konukoglu, and L.~V. Gool, ``Exploring
  cross-image pixel contrast for semantic segmentation,'' in
  \emph{{ICCV}}.\hskip 1em plus 0.5em minus 0.4em\relax {IEEE}, 2021, pp.
  7283--7293.

\bibitem{zhao2021contrastive}
X.~Zhao, R.~Vemulapalli, P.~A. Mansfield, B.~Gong, B.~Green, L.~Shapira, and
  Y.~Wu, ``Contrastive learning for label efficient semantic segmentation,'' in
  \emph{{ICCV}}.\hskip 1em plus 0.5em minus 0.4em\relax {IEEE}, 2021, pp.
  10\,603--10\,613.

\bibitem{tang2021proposal}
P.~Tang, C.~Ramaiah, Y.~Wang, R.~Xu, and C.~Xiong, ``Proposal learning for
  semi-supervised object detection,'' in \emph{{WACV}}.\hskip 1em plus 0.5em
  minus 0.4em\relax {IEEE}, 2021, pp. 2290--2300.

\bibitem{wei2021aligning}
F.~Wei, Y.~Gao, Z.~Wu, H.~Hu, and S.~Lin, ``Aligning pretraining for detection
  via object-level contrastive learning,'' in \emph{NeurIPS}, 2021, pp.
  22\,682--22\,694.

\bibitem{xie2021detco}
E.~Xie, J.~Ding, W.~Wang, X.~Zhan, H.~Xu, P.~Sun, Z.~Li, and P.~Luo, ``Detco:
  Unsupervised contrastive learning for object detection,'' in
  \emph{{ICCV}}.\hskip 1em plus 0.5em minus 0.4em\relax {IEEE}, 2021, pp.
  8372--8381.

\bibitem{gutmann2010noise}
M.~Gutmann and A.~Hyv{\"{a}}rinen, ``Noise-contrastive estimation: {A} new
  estimation principle for unnormalized statistical models,'' in
  \emph{{AISTATS}}, ser. {JMLR} Proceedings, vol.~9.\hskip 1em plus 0.5em minus
  0.4em\relax JMLR.org, 2010, pp. 297--304.

\bibitem{li2022exploring}
J.~Li, T.~Yang, W.~Ji, J.~Wang, and L.~Cheng, ``Exploring denoised cross-video
  contrast for weakly-supervised temporal action localization,'' in
  \emph{{CVPR}}.\hskip 1em plus 0.5em minus 0.4em\relax {IEEE}, 2022, pp.
  19\,882--19\,892.

\bibitem{zhang2021cola}
C.~Zhang, M.~Cao, D.~Yang, J.~Chen, and Y.~Zou, ``Cola: Weakly-supervised
  temporal action localization with snippet contrastive learning,'' in
  \emph{{CVPR}}.\hskip 1em plus 0.5em minus 0.4em\relax Computer Vision
  Foundation / {IEEE}, 2021, pp. 16\,010--16\,019.

\bibitem{paul2007emotions}
P.~\vspace{0mm} Ekman, ``Emotions revealed: recognizing faces and feelings to
  improve communication and emotional life,'' \emph{NY: OWL Books}, 2007.

\bibitem{wedel2009improved}
A.~Wedel, T.~Pock, C.~Zach, H.~Bischof, and D.~Cremers, ``An improved algorithm
  for tv-\emph{L} \({}^{\mbox{1}}\) optical flow,'' in \emph{Statistical and
  Geometrical Approaches to Visual Motion Analysis}, ser. Lecture Notes in
  Computer Science.\hskip 1em plus 0.5em minus 0.4em\relax Springer, 2008, vol.
  5604, pp. 23--45.

\bibitem{shou2018autoloc}
Z.~Shou, H.~Gao, L.~Zhang, K.~Miyazawa, and S.~Chang, ``Autoloc:
  Weakly-supervised temporal action localization in untrimmed videos,'' in
  \emph{{ECCV} {(16)}}, ser. Lecture Notes in Computer Science, vol.
  11220.\hskip 1em plus 0.5em minus 0.4em\relax Springer, 2018, pp. 162--179.

\bibitem{xu2016heterogeneous}
B.~Xu, Y.~Fu, Y.~Jiang, B.~Li, and L.~Sigal, ``Heterogeneous knowledge transfer
  in video emotion recognition, attribution and summarization,'' \emph{{IEEE}
  Trans. Affect. Comput.}, vol.~9, no.~2, pp. 255--270, 2018.

\bibitem{Shrivastava2016training}
A.~Shrivastava, A.~Gupta, and R.~B. Girshick, ``Training region-based object
  detectors with online hard example mining,'' in \emph{{CVPR}}.\hskip 1em plus
  0.5em minus 0.4em\relax {IEEE} Computer Society, 2016, pp. 761--769.

\bibitem{Girshick2012fast}
R.~B. Girshick, ``Fast {R-CNN},'' in \emph{{ICCV}}.\hskip 1em plus 0.5em minus
  0.4em\relax {IEEE} Computer Society, 2015, pp. 1440--1448.

\bibitem{see2019megc}
J.~See, M.~H. Yap, J.~Li, X.~Hong, and S.~Wang, ``{MEGC} 2019 - the second
  facial micro-expressions grand challenge,'' in \emph{{FG}}.\hskip 1em plus
  0.5em minus 0.4em\relax {IEEE}, 2019, pp. 1--5.

\bibitem{li2022megc2022}
J.~Li, M.~H. Yap, W.~Cheng, J.~See, X.~Hong, X.~Li, S.~Wang, A.~K. Davison,
  Y.~Li, and Z.~Dong, ``{MEGC2022:} {ACM} multimedia 2022 micro-expression
  grand challenge,'' in \emph{{ACM} Multimedia}.\hskip 1em plus 0.5em minus
  0.4em\relax {ACM}, 2022, pp. 7170--7174.

\bibitem{kingma2014adam}
D.~P. Kingma and J.~Ba, ``Adam: {A} method for stochastic optimization,'' in
  \emph{{ICLR} (Poster)}, 2015.

\bibitem{neubeck2006efficient}
A.~Neubeck and L.~V. Gool, ``Efficient non-maximum suppression,'' in
  \emph{{ICPR} {(3)}}.\hskip 1em plus 0.5em minus 0.4em\relax {IEEE} Computer
  Society, 2006, pp. 850--855.

\bibitem{wang2021understanding}
F.~Wang and H.~Liu, ``Understanding the behaviour of contrastive loss,'' in
  \emph{{CVPR}}.\hskip 1em plus 0.5em minus 0.4em\relax Computer Vision
  Foundation / {IEEE}, 2021, pp. 2495--2504.

\bibitem{he2020spotting}
Y.~He, S.~Wang, J.~Li, and M.~H. Yap, ``Spotting macro-and micro-expression
  intervals in long video sequences,'' in \emph{{FG}}.\hskip 1em plus 0.5em
  minus 0.4em\relax {IEEE}, 2020, pp. 742--748.

\bibitem{zhang2020spatio}
L.~Zhang, J.~Li, S.~Wang, X.~Duan, W.~Yan, H.~Xie, and S.~Huang,
  ``Spatio-temporal fusion for macro- and micro-expression spotting in long
  video sequences,'' in \emph{{FG}}.\hskip 1em plus 0.5em minus 0.4em\relax
  {IEEE}, 2020, pp. 734--741.

\bibitem{yuhong2021research}
Y.~He, ``Research on micro-expression spotting method based on optical flow
  features,'' in \emph{{ACM} Multimedia}.\hskip 1em plus 0.5em minus
  0.4em\relax {ACM}, 2021, pp. 4803--4807.

\bibitem{liong2022mtsn}
G.~B. Liong, S.~Liong, J.~See, and C.~Chee{-}Seng, ``{MTSN}: {A} multi-temporal
  stream network for spotting facial macro-and micro-expression with hard and
  soft pseudo-labels,'' in \emph{Proceedings of the 2nd Workshop on Facial
  Micro-Expression: Advanced Techniques for Multi-Modal Facial Expression
  Analysis}, 2022, pp. 3--10.

\bibitem{zhao2022rethink}
Y.~Zhao, X.~Tong, Z.~Zhu, J.~Sheng, L.~Dai, L.~Xu, X.~Xia, Y.~Jiang, and J.~Li,
  ``Rethinking optical flow methods for micro-expression spotting,'' in
  \emph{{ACM} Multimedia}.\hskip 1em plus 0.5em minus 0.4em\relax {ACM}, 2022,
  pp. 7175--7179.

\bibitem{gao2022fine}
J.~Gao, M.~Chen, and C.~Xu, ``Fine-grained temporal contrastive learning for
  weakly-supervised temporal action localization,'' in \emph{{CVPR}}.\hskip 1em
  plus 0.5em minus 0.4em\relax {IEEE}, 2022, pp. 19\,967--19\,977.

\end{thebibliography}

%

\begin{IEEEbiography}[{\includegraphics[width=1in,height=1.25in,clip,keepaspectratio]{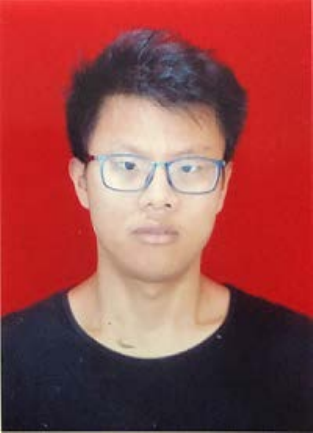}}]
  {Wang-Wang Yu} received the M.S. degree in biomedical engineering from University of Electronic Science and Technology of China (UESTC) in 2020. 
  He is now pursuing his Ph.D. degree in UESTC. His research interests include video understanding, emotional analysis, weakly supervised learning.
\end{IEEEbiography}

\begin{IEEEbiography}[{\includegraphics[width=1in,height=1.25in,clip,keepaspectratio]{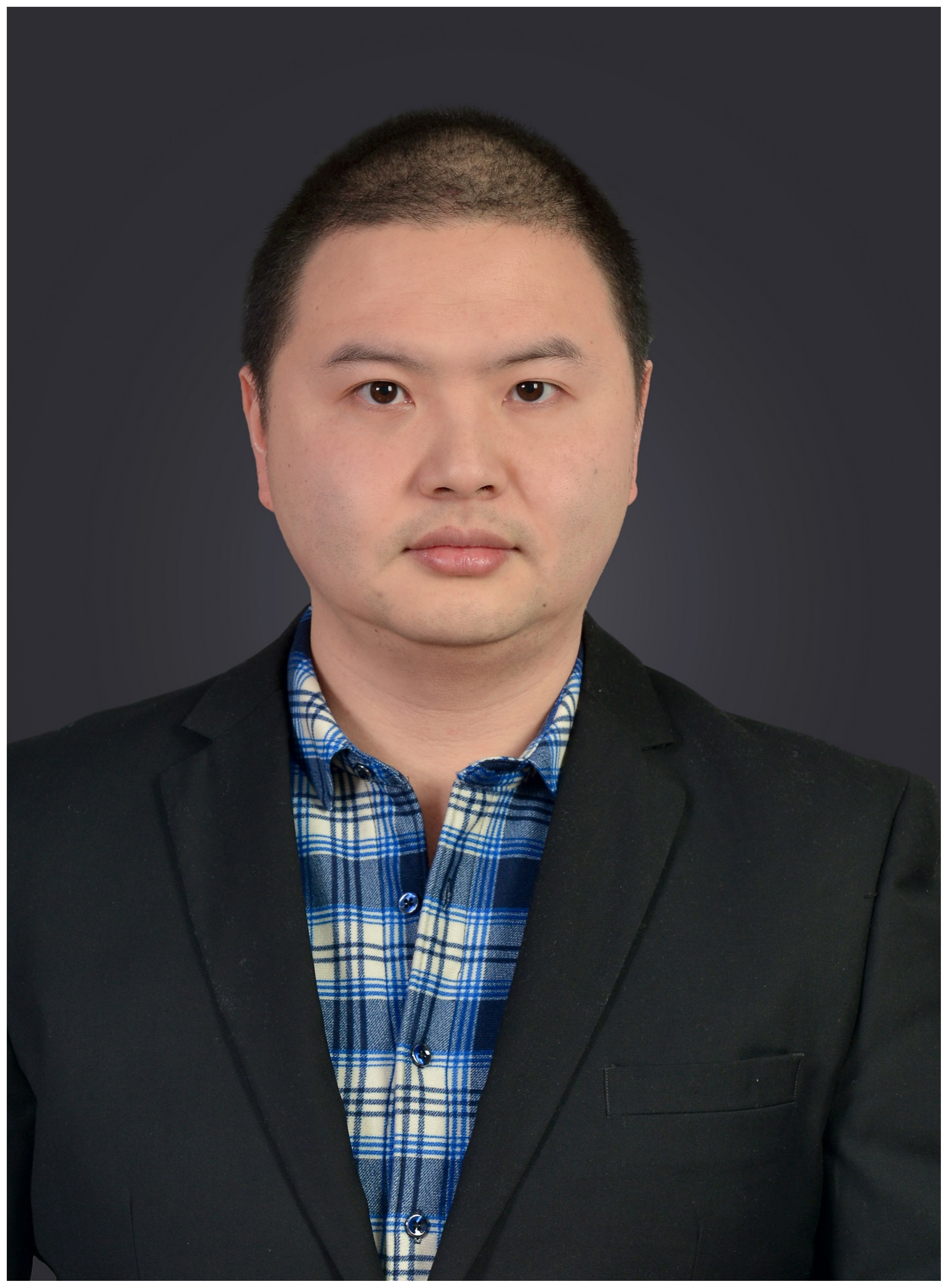}}]
  {Xian-Shi Zhang} Xian-Shi Zhang received the Ph.D. degree in biomedical engineering from the University of Electronic Science and Technology of China (UESTC), 
  Chengdu, China, in 2017. He is currently an Assistant Research Professor with the MOE Key Laboratory for Neuroinformation, School of Life Science and Technology, 
  UESTC. His research interests include visual mechanism modeling and bio-inspired computer vision.
\end{IEEEbiography}

\begin{IEEEbiography}[{\includegraphics[width=1in,height=1.25in,clip,keepaspectratio]{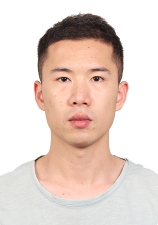}}]
  {Fu-Ya Luo} received the B.S. degree in biomedical engineering from University of Electronic Science and Technology of China (UESTC), in 2015. 
  He is now pursuing his Ph.D. degree in UESTC. His research interests include scene understanding, brain-inspired computer vision, 
  weakly supervised learning, and image-to-image translation.
\end{IEEEbiography}

\begin{IEEEbiography}[{\includegraphics[width=1in,height=1.25in,clip,keepaspectratio]{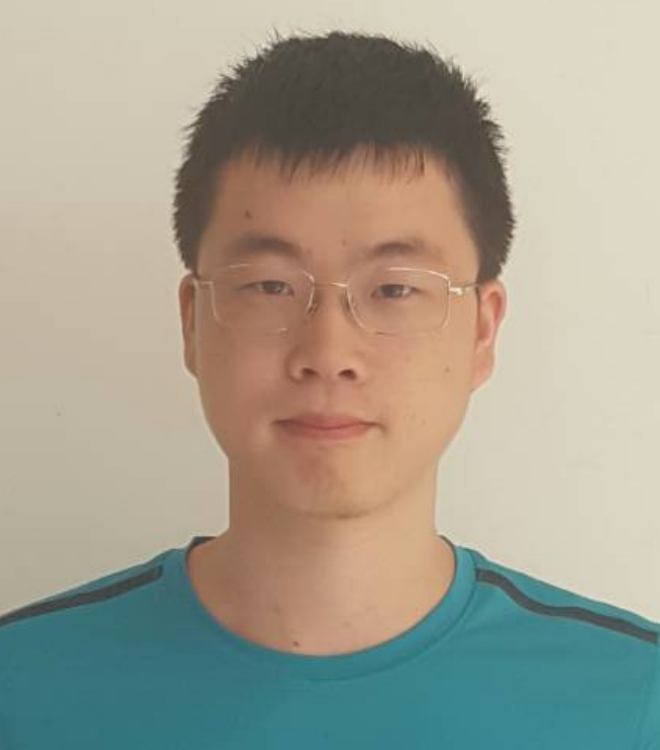}}]
  {Yijun Cao} received the M.S. degree from the
  College of Electric and Information Engineering,
  Guangxi University of Science and Technology. Currently, he is working toward the Ph.D. degree with
  the School of Life Science and Technology,
  University of Electronic Science and Technology of
  China (UESTC). His area of research is visual SLAM and navigation.
\end{IEEEbiography}

\begin{IEEEbiography}[{\includegraphics[width=1in,height=1.25in,clip,keepaspectratio]{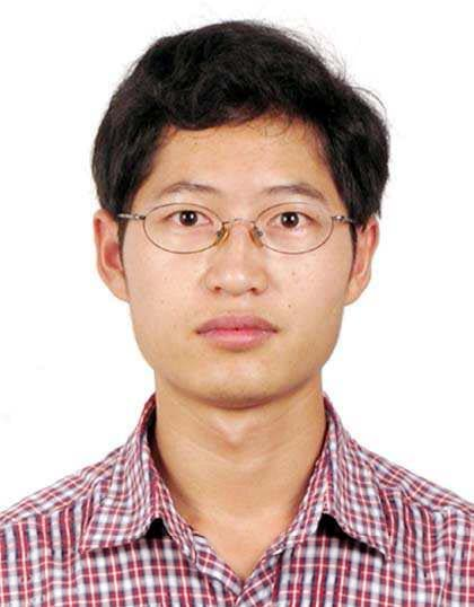}}]
  {Kai-Fu Yang} received the Ph.D. degree in biomedical engineering from the University of Electronic Science and 
  Technology of China (UESTC), Chengdu, China, in 2016. He is currently an associate research professor with 
  the MOE Key Lab for Neuroinformation, School of Life Science and Technology, UESTC, Chengdu, China. His research 
  interests include cognitive computing and brain-inspired computer vision.
\end{IEEEbiography}

\begin{IEEEbiography}[{\includegraphics[width=1in,height=1.25in,clip,keepaspectratio]{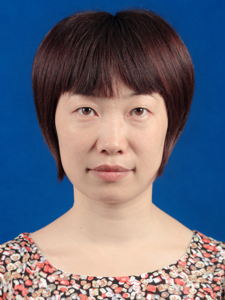}}]
  {Hong-Mei Yan} received the Ph.D. degree in biomedical engineering from Chongqing University in 2003. She is now a Professor 
  with the MOE Key Laboratory for Neuroinformation, University of Electronic Science and Technology of China, Chengdu, China. 
  Her research interests include visual cognition, visual attention, visual encoding and decoding.
\end{IEEEbiography}

\begin{IEEEbiography}[{\includegraphics[width=1in,height=1.25in,clip,keepaspectratio]{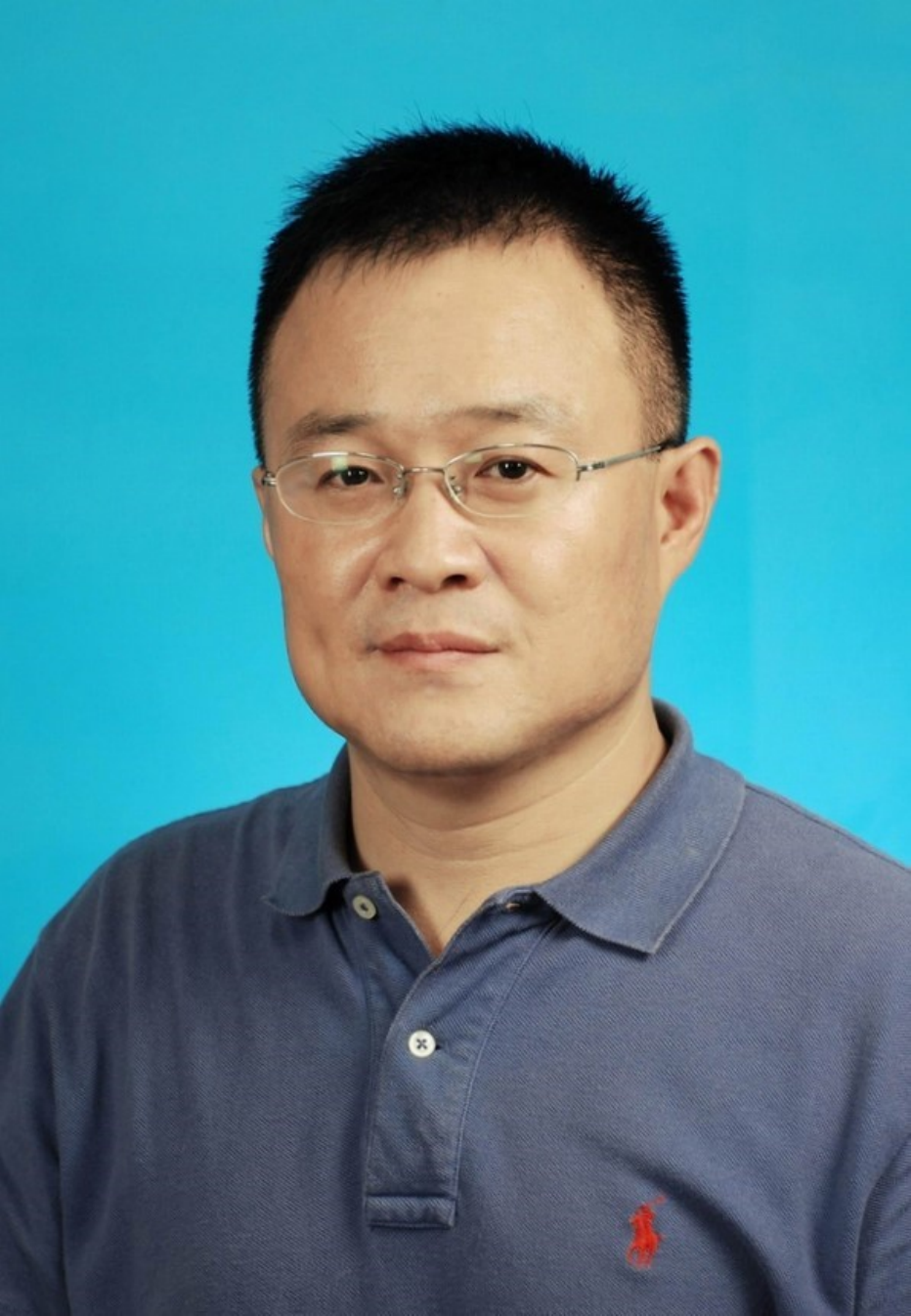}}]
  {Yong-Jie Li} (Senior Member, IEEE) received the Ph.D. degree in biomedical engineering from
  UESTC, in 2004. He is currently a Professor with the Key Laboratory for NeuroInformation of Ministry of
  Education, School of Life Science and Technology, University of Electronic Science and Technology of
  China. His research focuses on building of biologically inspired computational models of visual perception and 
  the applications in image processing and computer vision.
\end{IEEEbiography}



\end{document}